\documentclass[a4paper,fleqn]{cas-dc}

\usepackage{amsmath,amssymb,amsfonts}
\usepackage{algorithmic}
\usepackage{algorithm}
\usepackage{array}
\usepackage{placeins}
\usepackage{verbatim}
\usepackage{cite}
\usepackage{xcolor}
\usepackage{booktabs}
\usepackage{multirow}
\usepackage{etoolbox} 
\AtBeginEnvironment{algorithmic}{\linespread{1.1}\selectfont}
\usepackage{hyperref}

\usepackage[numbers]{natbib}

\begin{document}
\let\WriteBookmarks\relax
\def\floatpagepagefraction{1}
\def\textpagefraction{.001}
\shorttitle{STAIL}
\shortauthors{Songpan Gao et~al.}

\title [mode = title]{STAIL: Semantic Text-Anchored Incremental Learning for Medical Imaging via Large Language Models}   

\fntext[fn1]{Songpan Gao and Yajie Zhang contributed equally to this work.}
\fntext[fn2]{\url{https://github.com/Gao-leon/STAIL}}
\cortext[cor1]{Corresponding author: Zhi-An Huang (e-mail: huang.za@cityu-dg.edu.cn)}

\author[1]{Songpan Gao}
\ead{72403489@cityu-dg.edu.cn}
\credit{Conceptualization, Methodology, Software, Data curation, Validation, Investigation, Visualization, Writing - original draft}

\author[2]{Yajie Zhang}
\ead{yajie.zhang@connect.polyu.hk}
\credit{Conceptualization, Methodology, Software, Investigation, Formal analysis, Writing - review $\&$ editing}

\author[3,8]{Guanxing Chen}
\ead{guanxing.chen@cityu-dg.edu.cn}
\credit{Investigation, Validation, Writing - review $\&$ editing}

\author[3]{Jiayu Qian}
\ead{qjyariozzz@gmail.com}
\credit{Writing - review $\&$ editing, Methodology, Data curation}

\author[4]{Zhenzhen Liu}
\ead{liuzhenzhen@gzzoc.com}
\credit{Validation, Visualization, Data curation}

\author[5]{Shijun Li}
\ead{shijunli07@yeah.net}
\credit{Validation, Investigation}

\author[6]{Xiaowei Zhu}
\ead{xiaowei.zhu@cityu.edu.hk}
\credit{Validation, Investigation}

\author[2]{Yao Hu}
\ead{echo-yao.hu@polyu.edu.hk}
\credit{Writing - review $\&$ editing, Validation, Formal analysis}

\author[2]{Kay Chen Tan}
\ead{kctan@polyu.edu.hk}
\credit{Formal analysis, Resources, Funding acquisition, Supervision}

\author[7]{Yu-An Huang}
\ead{yuanhuang@nwpu.edu.cn}
\credit{Formal analysis, Writing - review $\&$ editing, Resources, Funding acquisition}

\author[8]{Shiqi Wang}
\ead{shiqwang@cityu.edu.hk}
\credit{Investigation, Validation, Writing - review $\&$ editing}

\author[3,8]{Zhi-An Huang}
\ead{huang.za@cityu-dg.edu.cn}
\credit{Supervision, Formal analysis, Methodology, Funding acquisition, Project administration, Resources, Writing - review $\&$ editing, Validation, Conceptualization}

\affiliation[1]{
organization={Department of Data Science, City University of Hong Kong (Dongguan)},
city={Dongguan},
postcode={523000},
country={China}
}

\affiliation[2]{
organization={Department of Data Science and Artificial Intelligence, The Hong Kong Polytechnic University},
city={Hong Kong},
country={Hong Kong SAR, China}
}

\affiliation[3]{
organization={Department of Computer Science, City University of Hong Kong (Dongguan)},
city={Dongguan},
postcode={523000},
country={China}
}

\affiliation[4]{
organization={State Key Laboratory of Ophthalmology, Zhongshan Ophthalmic Center, Sun Yat-sen University, Guangdong Provincial Key Laboratory of Ophthalmology and Visual Science},
city={Guangzhou},
postcode={510060},
country={China}
}

\affiliation[5]{
organization={Department of Radiology, the First Medical Center, Chinese PLA General Hospital},
city={Beijing},
postcode={100853},
country={China}
}

\affiliation[6]{
organization={Department of Neuroscience, City University of Hong Kong},
city={Hong Kong},
country={Hong Kong SAR, China}
}

\affiliation[7]{
organization={School of Computer Science, Northwestern Polytechnical University},
city={Xi'an},
postcode={710000},
country={China}
}

\affiliation[8]{
organization={Department of Computer Science, City University of Hong Kong},
city={Hong Kong},
country={Hong Kong SAR, China}
}

\begin{abstract}
Deep learning models applied to medical image analysis suffer from severe catastrophic forgetting when continually adapting to new clinical tasks in dynamic environments. Mainstream incremental learning methods typically mitigate this by rehearsing raw historical images. However, this pixel-level rehearsal incurs significant storage overhead, raises privacy concerns, and fails to adequately capture the true data distribution with sparse exemplars. Inspired by human cognitive mechanisms, we propose a novel framework termed Semantic Text-Anchored Incremental Learning (STAIL) for sequential clinical tasks. To overcome the rehearsal bottleneck, STAIL introduces an asymmetric semantic consolidation buffer (SCB). By  incorporating a minimal set of image anchors and extensive textual descriptions, the SCB enables dense semantic reconstruction of old tasks at a minimal storage cost. Furthermore, we design an LLM-derived Semantic Anchoring Mechanism (LSAM) that leverages the stable semantic space of frozen large language models as developmental priors. This mechanism explicitly anchors evolving visual features to textual representations, guiding and constraining plasticity and stability at both macroscopic and microscopic levels. Extensive experiments across three heterogeneous medical datasets, covering fundus, ultrasound, and X-ray imaging, demonstrate that STAIL acts as a highly effective plug-and-play module. It comprehensively enhances the performance of various existing baselines, achieving average gains of 2.24\% in AAA-AUC for sustained performance and 3.55\% in BWT-AUC for reduced forgetting. Code is available.
\end{abstract}

\begin{graphicalabstract}
\includegraphics[width=0.9\textwidth]{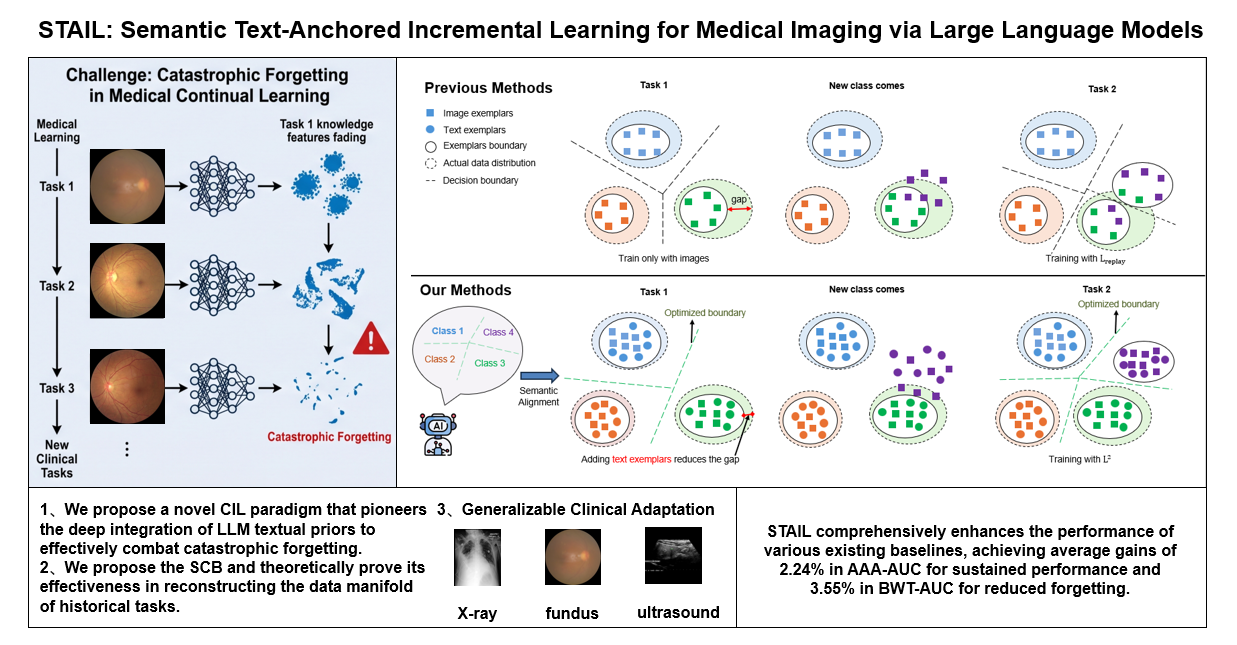}
\end{graphicalabstract}

\begin{highlights}
 \item We propose a novel CIL paradigm that pioneers the deep integration of LLM textual priors to effectively combat catastrophic forgetting.
\item We propose the SCB and theoretically prove its effectiveness in reconstructing the data manifold of historical tasks.
\item We provide deep mechanistic insights through extensive experimental analysis and feature visualization, explicitly revealing how textual priors help stabilize the visual feature space and optimize class decision boundaries.
\item We validate the plug-and-play versatility of the STAIL framework, demonstrating its ability to consistently and significantly enhance the performance of various mainstream incremental learning baselines across multiple heterogeneous medical datasets.
\end{highlights}


\maketitle

\section{Introduction}
Deep learning has significantly advanced medical data analysis~\cite{zhang2025anti,zhang2025causalmixnet,huang2026scbit}, yet most state-of-the-art (SOTA) models operate under a static learning paradigm. These models are typically trained on fixed datasets, failing to adapt to the dynamic nature of clinical practice where new diseases, evolving imaging protocols, and expanding diagnostic tasks frequently emerge. When updated with new data, traditional neural networks suffer from catastrophic forgetting~\cite{mccloskey1989catastrophic}, where newly acquired information erases previously learned knowledge, necessitating costly retraining from scratch. To address this, incremental learning has emerged as a pivotal paradigm in medical image analysis\cite{kirkpatrick2017overcoming}, aiming to seamlessly accumulate new knowledge while preserving historical expertise.

Existing efforts in Class-Incremental Learning (CIL) have primarily followed three technical trajectories: rehearsal-based methods~\cite{rebuffi2017icarl,10888025}, regularization-based strategies~\cite{kirkpatrick2017overcoming,zenke2017continual}, architecture-based approaches~\cite{wang2025enhancing,ayromlou2024ccsi,bayasi2025biaspruner}. Regularization-based strategies mitigate forgetting by constraining parameter updates, but often struggle with the rigidity-plasticity dilemma~\cite{wang2025dual}. Architecture-based approaches isolate interference by allocating task-specific parameters, yet face challenges with unbounded network expansion. Rehearsal-based methods, though highly effective by maintaining a representative buffer of past samples, incur significant storage overhead and privacy risks\cite{wang2024rehearsal,wang2025rethinking} when buffering raw medical images, and fail to filter redundant noise\cite{kumari2025continual}. In general, traditional CIL approaches predominantly focus on mathematical optimization or structural partitioning within a static framework. Consequently, they lack the semantic abstraction and cognitive flexibility required for complex clinical environments.

To overcome the limitations of these static frameworks and achieve true cognitive flexibility, we draw inspiration from biological learning systems. Specifically, we identify two fundamental yet underexplored discrepancies between current medical CIL and human cognition that motivate our work. The first discrepancy concerns memory representation. Neuroscientific evidence demonstrates that the human brain utilizes a sophisticated memory consolidation system that stores highly compressed, abstract semantic representations rather than verbatim sensory recordings~\cite{mcclelland1995there, kumaran2016learning}. The second discrepancy concerns knowledge initialization. Human learning is a cumulative, developmental process rooted in an extensive infant-stage observational period that establishes a rich foundation of prior world knowledge~\cite{gilboa2017neurobiology,smith2005development,lake2017building}. This developmental initialization phase provides the critical inductive bias to prevent over-fitting to early tasks and facilitates the graceful integration of new information. However, current CIL models start from scratch or narrow domain-specific pre-training.

In the realm of cognitive AI, natural language and large language models (LLMs) offer a compelling solution to both problems. Language is inherently a condensed, structured medium for information abstraction that mirrors semantic consolidation\cite{mirolli2009language}, and LLMs provide that foundational knowledge base through their pre-training. The key insight is that frozen LLM embeddings remain stationary and invariant to visual learning, making them ideal stable anchors to guide and constrain feature evolution across incremental tasks.

Motivated by this alignment, our biologically inspired framework, termed Semantic Text-Anchored Incremental Learning (STAIL), is engineered to facilitate dynamic medical diagnosis. STAIL leverages textual priors to emulate biological learning dynamics, utilizing text as a compressed memory format to bypass the pixel-level bottleneck, while exploiting the frozen semantic space of LLMs to safeguard the continuous accumulation of medical expertise. Specifically, STAIL addresses catastrophic forgetting through two synergistic components:
(1) Semantic consolidation buffer (SCB): To overcome the storage and privacy limitations of traditional rehearsal, the SCB employs an asymmetric design that stores a minimal set of core image anchors alongside a vast set of textual exemplars. This allows for a dense semantic reconstruction of historical task distributions~\cite{cheng2025distribution} at a negligible storage cost. (2) LLM-derived Semantic Anchoring Mechanism (LSAM): This module uses frozen LLM textual priors as stationary anchors to regulate visual feature evolution. By enforcing hierarchical constraints, including macroscopic semantic anchoring and microscopic contrastive refinement, the LSAM effectively balances feature space plasticity for novel tasks with structural stability for historical knowledge. The main contributions of this paper are as follows:
\begin{itemize}
    \item We propose a novel CIL paradigm that pioneers the deep integration of LLM textual priors to effectively combat catastrophic forgetting.
    
    \item We propose the SCB and theoretically prove its effectiveness in reconstructing the data manifold of historical tasks.
    
    \item We provide deep mechanistic insights through extensive experimental analysis and feature visualization, explicitly revealing how textual priors help stabilize the visual feature space and optimize class decision boundaries.
    
    \item We validate the plug-and-play versatility of the STAIL framework, demonstrating its ability to consistently and significantly enhance the performance of various mainstream incremental learning baselines across multiple heterogeneous medical datasets.
\end{itemize}

\section{Related Work}
\label{cap:relatedwork}
This section reviews two pivotal streams of research: the evolution of CIL and the integration of LLMs into incremental paradigms. We specifically focus on identifying the inherent limitations of pixel-level rehearsal and the constraints of current prompt-learning methods, highlighting the technical gaps that our bio-inspired mechanisms aim to bridge.
\subsection{Class-incremental Learning and Rehearsal Limitations}
CIL aims to empower deep models to sequential clinical tasks without catastrophic forgetting. Existing approaches predominantly fall into three categories: regularization-based, architecture-based, and rehearsal-based methods. While regularization strategies~\cite{li2017learning,li2025progressive} like EWC~\cite{kirkpatrick2017overcoming} bypass the need for data storage, they are often constrained by the rigidity-plasticity dilemma. Dynamic architectures~\cite{wang2022foster,wang2022beef,bayasi2024gc,wang2025cross} including DER~\cite{yan2021dynamically} and MEMO~\cite{zhoumodel} tackle forgetting through structural expansion, yet they frequently suffer from unbounded parameter proliferation, limiting their scalability in resource-constrained medical devices.
Consequently, rehearsal-based methods~\cite{thandiackal2024multi}, which maintain a representative buffer of historical exemplars, remain the most robust and widely adopted CIL paradigms. 

Foundational works in this domain, such as iCaRL~\cite{rebuffi2017icarl}, established the baseline by combining knowledge distillation with a nearest-mean-of-exemplars classifier. Subsequent advancements have largely focused on optimizing exemplars utilization. For instance, weight aligning (WA)~\cite{zhao2020maintaining} approach addresses catastrophic forgetting by correcting the biased weights of the final classifier to ensure representational fairness between old and new classes. Furthermore, contemporary hybrid strategies like TagFex~\cite{zheng2025task} introduce dual-branch frameworks, utilizing stable, task-agnostic features to guide task-specific expansion without corrupting consolidated representations. Despite these algorithmic refinements, a fundamental bottleneck persists: these methods rely heavily on buffering raw medical images. As highlighted in our introduction, this conventional pixel-level rehearsal imposes significant storage overhead, exacerbates patient privacy vulnerabilities, and inadvertently preserves scanner-specific artifacts that hinder generalized semantic consolidation. 
\begin{figure*}[t]
    \centering
    \includegraphics[width=1\linewidth]{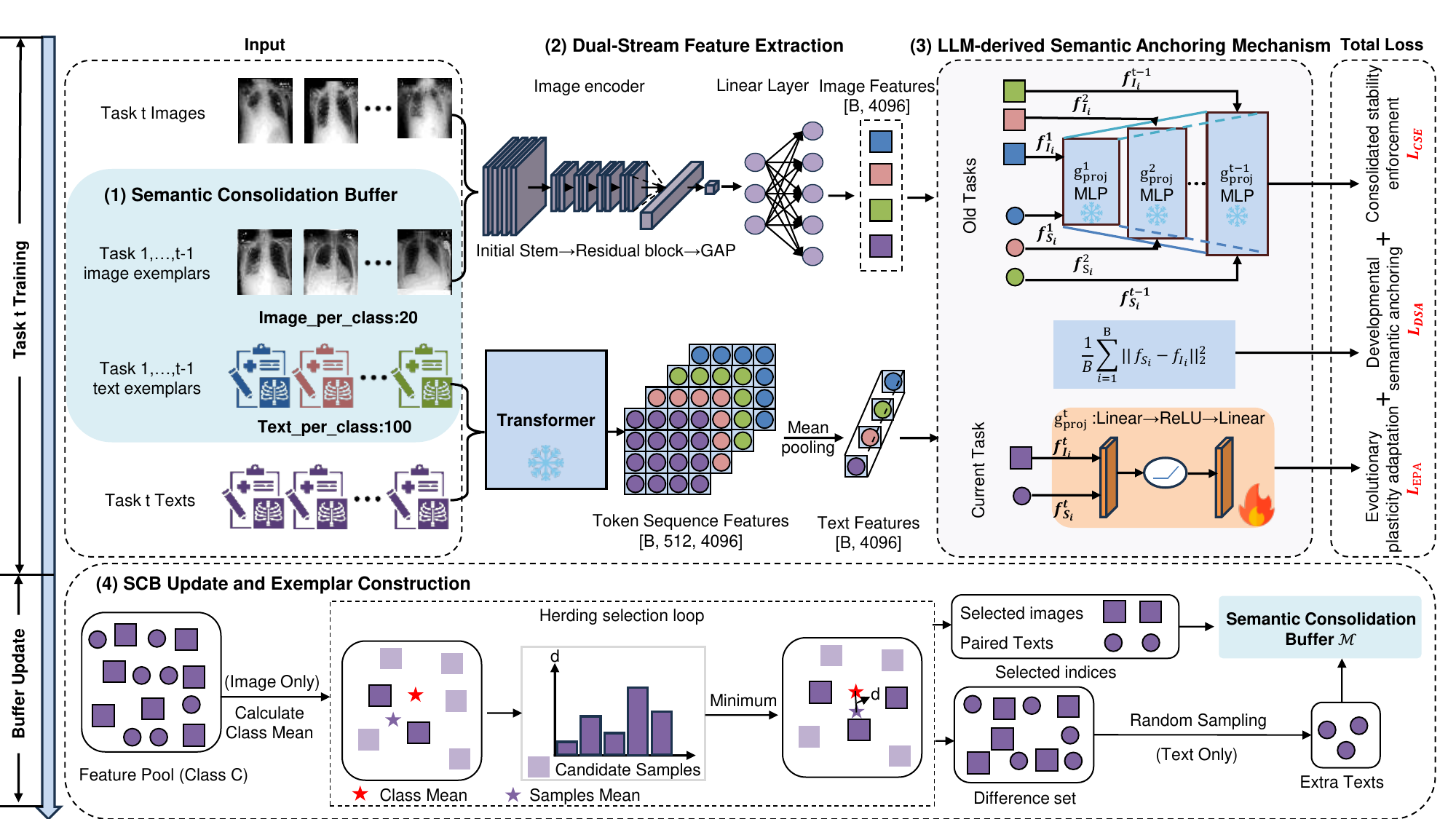}
    \caption{Schematic illustration of the proposed STAIL framework. The top panel depicts the training process for task $t$, where the model accepts current task data combined with historical asymmetric exemplars retrieved from the SCB as input. Following dual-stream feature extraction via a trainable visual encoder and a frozen LLM, the LSAM regularizes the feature space through three losses: $\mathcal{L}_{DSA}$, $\mathcal{L}_{EPA}$, and $\mathcal{L}_{CSE}$. The bottom panel details the post-training SCB update and exemplar construction phase.}
    \label{fig:task t}
\end{figure*}

\subsection{Large Language Models and Vision-Language Models in CIL}
The emergence of vision-language models (VLMs)~\cite{radford2021CLIP,qian2025cpsr} has triggered a new wave of CIL research leveraging pre-trained semantic priors. The prompt-based learning approaches, such as L2P~\cite{wang2022L2P} and DualPrompt~\cite{wang2022dualprompt}, preserve the knowledge of pre-trained visual encoders by freezing the backbone and optimizing task-adaptive prompt parameters. These methods typically construct textual prompts based on class-level semantic descriptions, where the text representation mainly serves as a category-level prototype for guiding visual recognition.

However, class-level textual prompts provide only coarse-grained semantic information and may overlook substantial intra-class variations that are critical in medical scenarios. In contrast, image-level clinical descriptions, such as radiology reports, contain instance-specific semantic cues, including anatomical locations, disease severity, and heterogeneous manifestations. Recent medical vision-language studies~\cite{boecking2022making,BUI2026104235} have demonstrated that image-level text can capture richer fine-grained medical semantics beyond class labels. Different from existing prompt-based CIL methods that mainly exploit class-level textual prompts with frozen visual backbones, our framework utilizes image-level clinical text anchors as semantic memory to regularize the evolution of the visual representation. By anchoring continuously learned visual features to stationary LLM derived semantic embeddings, STAIL enables the visual encoder to adapt to emerging medical distributions while preserving clinically meaningful semantic structures.

\section{METHODOLOGY}
\label{cap:method}
In this section, we propose STAIL, a bio-inspired framework that leverages LLM textual priors as stationary anchors for continuous medical knowledge acquisition. As illustrated in Fig. \ref{fig:task t}, our approach addresses the inherent bottlenecks of traditional pixel-level rehearsal through two synergistic components: an asymmetric semantic consolidation buffer for enhanced memory efficiency, and an LLM-derived Semantic Anchoring Mechanism for robust feature regulation. The following subsections detail the problem formulation and proposed modules. The theoretical insights and pseudo-code of STAIL are provided in Appendix~\ref{Theoretical Insights} and~\ref{Pseudo-code}

\subsection{Problem Formulation: Traditional Rehearsal-Based CIL}
In the standard CIL setting, a model learns from a sequence of $K$ disjoint tasks, denoted as $\mathcal{T}=\{T_{1},T_{2},...,T_{K}\}$. At each task $t$, the model receives a dataset $\mathcal{D}_{t}=\{(I_{i}^{t},y_{i}^{t})\}_{i=1}^{N_{t}}$, where $I_{i}^{t}$ is a raw medical image, $y_{i}^{t}$ is its corresponding label, and $N_{t}$ represents the number of samples for the $t$-th task. The objective of traditional rehearsal-based methods is to minimize the empirical risk on the current task $T_{t}$ while preserving knowledge from previous tasks $\mathcal{T}_{1:t-1}$. Because these methods rely solely on visual retention, they maintain a strictly limited memory buffer $\mathcal{M}_{img}$ storing a small subset of raw images from previous tasks. The standard optimization objective is defined as:
\begin{equation}
\label{LossTraCIL}
    \min_{\theta} \mathbb{E}_{(I, y) \sim \mathcal{D}_t \cup \mathcal{M}_{img}} [\mathcal{L}_{Base}^t(C_{Base}(g_I(I)), y)],
\end{equation}
where $\theta$ encapsulates the learnable parameters of the network, $g_I(\cdot)$ is the visual backbone, $C_{Base}(\cdot)$ is the classifier, and $\mathcal{L}_{Base}^t$ represents the mainstream CIL loss function, such as cross-entropy or distillation loss. By focusing strictly on risk minimization via a limited raw image buffer, this traditional formulation inevitably suffers from a pixel-level bottleneck, overlooking the cognitive benefits of semantic memory compression.

\subsection{Framework Overview}
Our STAIL framework introduces a semantically-anchored learning paradigm, which draws inspiration from human cognitive mechanisms, specifically, semantic memory consolidation and developmental initialization. 
Formally, we redefine the training dataset for the $t$-th task to incorporate the auxiliary textual modality: $\mathcal{D}_{t}=\{(I_{i}^{t},S_{i}^{t},y_{i}^{t})\}_{i=1}^{N_{t}}$, where $S_{i}^{t}$ is the text description(e.g., diagnostic reports). As illustrated in Fig. \ref{fig:task t}, the workflow of STAIL consists of four core phases:

\textbf{Semantic Consolidation Buffer (SCB at Input Phase)}. During the training of task $t$, the network processes the current task's data alongside historical exemplars retrieved from our SCB. Emulating the brain's efficient memory compression, the SCB breaks away from traditional image-heavy buffers by employing an asymmetric design: it utilizes a Herding strategy to  extract a minimal core of representative visual anchors, which are then complemented by a massive volume of highly compressed semantic descriptors (comprising both paired and randomly sampled auxiliary texts).

\textbf{Dual-Stream Feature Extraction (DSFE)}. The multimodal training inputs are routed through distinct pathways. Raw medical images are fed into a trainable visual encoder to generate evolving visual features. Concurrently, the auxiliary textual reports are passed through a frozen LLM to extract highly stable semantic features. This pre-trained LLM acts as a computational analogue to human developmental priors, providing a robust anchor of general world knowledge.

\textbf{LLM-derived Semantic Anchoring Mechanism (LSAM)}. To orchestrate visual feature evolution and mitigate catastrophic forgetting, the LSAM leverages frozen text priors to optimize the feature space via three complementary semantic regularization objectives: (1) Developmental semantic anchoring ($\mathcal{L}_{DSA}$) prevents macroscopic drift by tethering evolving visual features to the stable semantic space of the frozen LLM. (2) Evolutionary plasticity adaptation ($\mathcal{L}_{EPA}$) ensures plasticity for new tasks by using contrastive learning to enhance intra-class compactness and inter-class separability. (3) Consolidated stability enforcement ($\mathcal{L}_{CSE}$) preserves the structural integrity of old classes by utilizing historical projectors and SCB texts to replay past geometric constraints.

\textbf{SCB Update and Exemplar Construction}. Post-training, the SCB is updated to efficiently reconstruct the data manifold of old tasks at a minimal storage cost. To maintain the asymmetric budget, image exemplars are selected via a Herding strategy~\cite{welling2009herding} to retain a representative visual core. Concurrently, we combine the paired textual reports with an extensive collection of supplementary texts retrieved via random sampling from the difference set. This strategy effectively captures the broader semantic diversity that sparse visual anchors might miss.

\textbf{Overall Objective}. By integrating the cognitive constraints from both the asymmetric SCB and the LSAM, the total training objective for the STAIL framework at task $t$ is a comprehensive sum of the baseline loss and the three developmental regulation terms:
\begin{equation}
\label{totalloss}
\mathcal{L}^{t} = \mathcal{L}_{Base}^{t} + \lambda_{DSA}\mathcal{L}_{DSA}^{t} + \lambda_{EPA}\mathcal{L}_{EPA}^{t} + \lambda_{CSE}\mathcal{L}_{CSE}^{t},
\end{equation}
where $\lambda_{DSA}$, $\lambda_{EPA}$, and $\lambda_{CSE}$ are hyper-parameters balancing the loss components.
This formulation allows our framework to seamlessly empower any existing rehearsal-based CIL baseline without requiring structural modifications to their original optimization objectives.

\subsection{Semantic Consolidation Buffer and Dual-Stream Feature Extraction}
This section details the operationalization of SCB and DSFE of the STAIL framework. During training on task $t$, the model ingests not only the current task data $\mathcal{D}_{t}$ but also historical exemplars retrieved from the SCB, forming an integrated input stream $(I_i, S_i) \sim \mathcal{D}_{t} \cup \mathcal{M}$. 

The SCB, denoted as $\mathcal{M}$, maintains an asymmetric multimodal memory by storing a limited number of image exemplars $\mathcal{M}_{img}$ and a text memory $\mathcal{M}_{text}$ from previous tasks. Specifically, $\mathcal{M}_{text}$ consists of two complementary components: the paired clinical descriptions $\mathcal{E}_{pair}$ associated with the retained image exemplars and additional text descriptions $\mathcal{E}_{extra}$. Therefore, the text memory can be formulated as:
\begin{equation}
\mathcal{M}_{text}=\mathcal{E}_{pair}\cup\mathcal{E}_{extra}.
\end{equation}
This design achieves substantial storage efficiency while preserving rich semantic information for knowledge replay. The combined multimodal inputs are processed through two parallel streams as follows.

\textbf{Semantic Stream}. The text descriptions $S_i$ are passed through a frozen LLM text encoder $g_{llm}(\cdot)$ to extract high-quality, stationary semantic features $f_{S_i} = g_{llm}(S_i) \in \mathbb{R}^{D_{text}}$, where $D_{text}$ denotes semantic feature dimension.

\textbf{Visual Stream}. Raw medical images $I_i$ are fed into a trainable visual encoder $g_I(\cdot)$, for example ResNet-18~\cite{he2016deep}, to generate evolving visual features $g_I(I_i) \in \mathbb{R}^{D_{vis}}$, where $D_{vis}$ represents the visual feature dimension. To establish a cross-modal bridge between the trainable visual encoder $g_I(\cdot)$ and the frozen LLM text encoder $g_{llm}(\cdot)$, we introduce a learnable linear projection matrix $W_{proj} \in \mathbb{R}^{D_{vis} \times D_{text}}$ at the end of the visual stream, explicitly mapping the visual representations into the shared semantic space, yielding $f_{I_i}=W_{proj}g_I(I_i)  \in \mathbb{R}^{D_{text}}$.

By processing the multimodal inputs in parallel, we establish the fundamental feature space required for the subsequent developmental calibration, treating the frozen LLM outputs as stable developmental priors.

\subsection{LLM-derived Semantic Anchoring Mechanism}
This core optimization engine calibrates the feature space via three constraints, systematically balancing macroscopic stability, microscopic plasticity, and historical structural integrity.

\textbf{Developmental Semantic Anchoring ($\mathcal{L}_{DSA}$)}. To prevent the macroscopic drift of visual representations when the network adapts to new clinical tasks, we utilize the frozen LLM embedding as stable semantic anchors. Given a batch of image-text feature representations ($f_{I_i}, f_{S_i}$), where $I_i \in D^{I}_t \cup M_{img}$ and $S_i \in D^{S}_t \cup \mathcal{E}_{pair}$, we employ a mean squared error (MSE) alignment loss to explicitly pull the projected visual features towards their corresponding text anchors, enforcing tight macroscopic coupling for both new and old classes:
\begin{equation}
    \mathcal{L}_{DSA}^{t}=\frac{1}{B}\sum_{i=1}^{B}||f_{S_i}-f_{I_i}||_2^2,
\end{equation}
where $B$ denotes the number of aligned image-text pairs in the batch.

\textbf{Evolutionary Plasticity Adaptation ($\mathcal{L}_{EPA}$)}. While the semantic anchoring ensures global stability, the model must simultaneously maintain microscopic plasticity to distinguish fine-grained nuances in newly arriving pathologies. Operating exclusively on the current task dataset $\mathcal{D}_t$, this module focuses on refining feature discriminability for the novel classes. For the $t$-th task, we construct a task-specific contrastive projection head $g_{proj}^t$, designed as a multi-layer perceptron (MLP), to map features into a normalized hypersphere. Given the current data $(I_i, S_i) \sim \mathcal{D}_t$, the module generates $L_2$-normalized projected features, denoted as $\hat{f}_{I_i} = Norm(g_{proj}^t(f_{I_i}))$ and $\hat{f}_{S_i} = Norm(g_{proj}^t(f_{S_i}))$. 
To enhance intra-class semantic compactness and inter-class separability within the shared projection space, we introduce a class-aware image-text contrastive objective. Specifically, given an image batch $\mathcal{B}_{I}=\{(I_i,y_i)\}_{i=1}^{B_I}$ and a text batch $\mathcal{B}_{S}=\{(S_j,y_j)\}_{j=1}^{B_S}$, where $B_I$ and $B_S$ represent their corresponding batch sizes, respectively, we define the positive text set for each image as: $P(i)=\{j|y_j=y_i\},$where $P(i)$ indicates the set of text samples in $\mathcal{B}_{S}$ that share the same class label with the image sample $I_i$. The EPA loss is formulated as:
\begin{equation}
\mathcal{L}_{EPA}^{t}
=
-\frac{1}{B_I}
\sum_{i=1}^{B_I}
\frac{1}{|P(i)|}
\sum_{j\in P(i)}
\log
\frac{
\exp(sim(\hat f_{I_i},\hat f_{S_j})/\tau)
}
{
\sum_{l=1}^{B_S}
\exp(sim(\hat f_{I_i},\hat f_{S_l})/\tau)
},
\end{equation}
where $sim(\cdot,\cdot)$ represents the cosine similarity, and $\tau$ is the temperature scalar controlling the penalty strength on hard negative samples. 

\textbf{Consolidated Stability Enforcement ($\mathcal{L}_{CSE}$)}. To directly combat catastrophic forgetting at a microscopic level, standard rehearsal utilizing only a classification loss is insufficient. The $\mathcal{L}_{CSE}$ addresses this by explicitly replaying the geometric constraints of past tasks, leveraging the model's past self to ensure that the structural integrity of old classes remains undistorted. Structurally, we freeze the task-specific projection heads trained during all previous incremental steps, denoted as $\{g_{proj}^{k}\}_{k=1}^{t-1}$. These frozen MLPs act as historical reference networks to simulate the feature distribution of task $k$ exactly as it was originally learned. Unlike the previous modules, the inputs for $\mathcal{L}_{CSE}$ are explicitly drawn from the SCB, consisting of historical image exemplars $\mathcal{M}_{img}$ and historical text memory $\mathcal{M}_{text}$. The text memory contains both paired clinical descriptions $\mathcal{E}_{pair}$ associated with retained images and supplementary text samples $\mathcal{E}_{extra}$. Specifically, $\mathcal{E}_{extra}$ provides additional semantic coverage beyond the limited visual memory. By introducing diverse textual descriptions of previous classes, it enriches the semantic support of historical categories, providing more comprehensive class-level constraints and reducing the risk of overfitting to a small number of stored image exemplars.

This stability loss enforces that the current model's representation of old data, when evaluated through the lens of these historical projectors, maintains the original contrastive geometric boundaries:
\begin{equation}
\begin{aligned}
\mathcal{L}_{CSE}^{t} &
= 
-\sum_{k=1}^{t-1}
\frac{1}{B_I}
\sum_{i=1}^{B_I}
\frac{1}{|P(i)|}
\sum_{j\in P(i)} \\
&
\log
\frac{
\exp(
sim(
g_{proj}^{k}(f_{I_i}),
g_{proj}^{k}(f_{S_j})
)/\tau
)
}
{
\sum_{l=1}^{B_S}
\exp(
sim(
g_{proj}^{k}(f_{I_i}),
g_{proj}^{k}(f_{S_l})
)/\tau
)
}.
\end{aligned}
\end{equation}
This formulation guarantees that the integration of new knowledge does not overwrite the topological consistency established in previous tasks.

\subsection{SCB Update and Exemplar Construction}
Upon the completion of the training phase for the $t$-th task, we update the SCB to incorporate the newly acquired knowledge. Let $\mathcal{C}_t$ represent the set of classes introduced in the current task $t$, and for each class $c \in \mathcal{C}_t$, let $\mathcal{D}_c^t$ denote its corresponding training dataset. We enforce an asymmetric budget allocating a significantly larger quota for text exemplars ($N_{text}$) than for image exemplars ($N_{img}$), such that $N_{text} \gg N_{img}$. For each new class $c \in \mathcal{C}_t$, the exemplar selection and buffer update process follow a two-stage mechanism aimed at maximizing information entropy.

\textbf{Visual Core Selection (Herding)}. First, to preserve the visual manifold's center, we employ a Herding strategy in the feature space. We compute the mean feature vector $\mu_c$ for class $c$ using the trained visual encoder. We then iteratively select a minimal set of image exemplars $\mathcal{E}_{img}^c$, that best approximate this mean. Crucially, for every selected image $I_k \in \mathcal{E}_{img}^c$, its paired text report $S_k\in \mathcal{E}_{pair}^c$ is synchronously retained. This step ensures a precise alignment between core visual features and their corresponding semantic descriptions.

\textbf{Semantic Expansion (Random Sampling)}. Subsequently, to capture the broader semantic diversity that sparse visual anchors might miss, we define a candidate pool $\mathcal{U}_{c}^t$ by calculating the difference set between the full dataset $\mathcal{D}_c^t$ and the selected core image-text pairs:
\begin{equation}
    \mathcal{U}_{c}^t = \mathcal{D}_c^t \setminus \{ (I, S) \mid I \in \mathcal{E}_{img}^c, S \in \mathcal{E}_{pair}^c \},
\end{equation}
where $\mathcal{D}_c^t$ denotes the complete set of training image-text pairs for class $c$ in the current task $t$. From this candidate pool $\mathcal{U}_{c}^t$, we apply a random sampling strategy to acquire a massive set of additional, purely textual reports $\mathcal{E}_{extra}^c$.

Finally, we update the global memory buffer $\mathcal{M}$ by uniting the selected sets from all classes in the current task. The image buffer $\mathcal{M}_{img}$ and text buffer $\mathcal{M}_{text}$ are formally updated as follows:
\begin{equation}
\begin{split}
\mathcal{M}_{img}  &\leftarrow \mathcal{M}_{img} \cup \bigcup_{c \in \mathcal{C}_t} \mathcal{E}_{img}^c, \\
\mathcal{M}_{text} &\leftarrow \mathcal{M}_{text} \cup \bigcup_{c \in \mathcal{C}_t} (\mathcal{E}_{pair}^c \cup \mathcal{E}_{extra}^c).
\end{split}
\end{equation}
This formulation ensures that for every class learned up to task $t$, the buffer maintains a dense semantic reconstruction of the original distribution, stably anchored by a sparse visual core.

\section{EXPERIMENTS}
To comprehensively evaluate the STAIL framework, we conducted extensive experiments on three heterogeneous medical datasets to verify its robustness across diverse modalities. This section systematically demonstrates its effectiveness in mitigating catastrophic forgetting by seamlessly integrating STAIL as a plug-and-play module into mainstream replay-based baselines and comparing it against state-of-the-art CIL methods.

\label{cap:experiment}
\subsection{Experimental Setups}
\subsubsection{Datasets and Protocols}
We conducted extensive experiments on three representative datasets, covering X-ray, ultrasound, and fundus photography.

\textbf{ODIR-5K~\cite{odir2019}}. For ophthalmic disease analysis, we employed the ODIR-5K (Ocular Disease Intelligent Recognition) dataset. It consists of color fundus photographs from 5,000 patients, covering both left and right eyes, accompanied by clinical diagnostic keywords. We organized the data into 8 distinct diagnostic categories. Notably, although the original dataset provided predefined training and testing splits, a portion of the training set and the entire testing set lacked corresponding text reports. Since our STAIL framework requires complete image-text pairs, we filtered the dataset to retain only samples with available textual descriptions. To prevent potential patient-level data leakage, we further ensured that images from the same patient were assigned exclusively to a single data split. Specifically, we verified the patient identifiers of all samples and confirmed that no patient appeared across the training, validation, and testing sets. The resulting dataset was partitioned into training, validation, and testing sets with a ratio of 70:15:15. We used half of the classes to train the initial model and evenly introduced two new classes at each subsequent incremental learning step.

\textbf{US-DATA~\cite{li2024ultrasound}}. We adopted a large-scale clinical ultrasound dataset, referred to as US-DATA, containing ultrasound images and report data from over 7,300 patients, to evaluate performance on fine-grained classification across different anatomical structures. The dataset comprises ultrasound images of 41 cluster categories across three major organs: liver (18 classes), thyroid (5 classes), and mammary (18 classes). We strictly followed the dataset's official predefined training, validation, and testing splits. We trained the initial model on 11 classes, and the remaining classes are grouped into incremental learning tasks of 6 classes each.

\textbf{MS-CXR~\cite{PhysioNet-ms-cxr-1.1.0}}. A curated subset of the MIMIC-CXR database verified by radiologists, consists of 1,162 high-quality image-text pairs spanning 8 major radiological findings. For MS-CXR, we followed the official patient-level split provided by the dataset. We used half of the classes to train the initial model and evenly introduce two new classes at each incremental learning step.

\begin{table*}[t]
\centering
\caption{Comparison of CIL Performance on Three Medical Datasets.}
\label{table:main_results}
\renewcommand{\arraystretch}{1.3} 

\setlength{\tabcolsep}{3pt}

\resizebox{\textwidth}{!}{%
\begin{tabular}{l|ccc|ccc|ccc}
\hline
\multirow{2}{*}{\textbf{Method}} & \multicolumn{3}{c|}{\textbf{ODIR-5K}} & \multicolumn{3}{c|}{\textbf{US-DATA}} & \multicolumn{3}{c}{\textbf{MS-CXR}} \\
\cline{2-10} 
 & Avg-AUC & AAA-AUC & BWT-AUC & Avg-AUC & AAA-AUC & BWT-AUC & Avg-AUC & AAA-AUC & BWT-AUC \\ 
\hline
EWC~\cite{kirkpatrick2017overcoming}& 64.13 $\pm$ 1.72 & 70.02 $\pm$ 1.32 & -20.38 $\pm$ 1.13 & 51.64 $\pm$ 3.47 & 65.20 $\pm$ 0.55 & -38.41 $\pm$ 3.87 & 59.08 $\pm$ 4.03 & 56.82 $\pm$ 2.94 & \textbf{3.19} $\pm$ \textbf{4.49} \\
\hline
DER~\cite{yan2021dynamically} & 70.52 $\pm$ 0.70 & 72.76 $\pm$ 1.08 & -6.89 $\pm$ 1.91 & 78.82 $\pm$ 0.87 & 79.36 $\pm$ 0.08 & -8.44 $\pm$ 1.05 & 57.79 $\pm$ 2.82 & 59.08 $\pm$ 1.60 & -1.43 $\pm$ 0.83 \\
\hline
MEMO~\cite{zhoumodel} & 67.05 $\pm$ 1.99 & 71.71 $\pm$ 0.96 & -14.91 $\pm$ 1.02 & 73.55 $\pm$ 2.36 & 74.81 $\pm$ 0.92 & -13.54 $\pm$ 2.85 & 54.33 $\pm$ 2.73 & 59.34 $\pm$ 1.34 & -5.69 $\pm$ 2.72 \\
\hline
L2P~\cite{wang2022L2P} & 69.85 $\pm$ 1.21 & 71.50 $\pm$ 2.55 & -6.84 $\pm$ 0.23 & 77.10 $\pm$ 1.81 & 78.89 $\pm$ 1.92 & -8.17 $\pm$ 0.49 & 59.75 $\pm$ 3.45 & 60.19 $\pm$ 0.45 & 0.55 $\pm$ 1.18 \\
\hline
DualPrompt~\cite{wang2022dualprompt} & 70.14 $\pm$ 1.93 & 72.30 $\pm$ 1.18 & -5.49 $\pm$ 1.18 & 76.49 $\pm$ 2.19 & 79.92 $\pm$ 0.72 & -7.85 $\pm$ 1.86 & 58.23 $\pm$ 4.65 & 60.06 $\pm$ 3.63 & 0.03 $\pm$ 0.62 \\
\hline
Replay  & 66.97 $\pm$ 3.67 & 71.81 $\pm$ 1.12 & -17.65 $\pm$ 6.35 & 70.64 $\pm$ 0.75 & 73.32 $\pm$ 0.18 & -17.30 $\pm$ 1.37 & 59.29 $\pm$ 3.01 & 59.36 $\pm$ 1.69 & -0.96 $\pm$ 2.64 \\
+STAIL(Ours)  & 67.60 $\pm$ 1.44\,\ensuremath{\uparrow_{\textcolor{blue}{\scriptsize 0.63}}}
 & 72.53 $\pm$ 1.03\,\ensuremath{\uparrow_{\textcolor{blue}{\scriptsize 0.72}}} & -14.01 $\pm$ 2.80\,\ensuremath{\uparrow_{\textcolor{blue}{\scriptsize 3.64}}} & 78.74 $\pm$ 0.48\,\ensuremath{\uparrow_{\textcolor{blue}{\scriptsize 8.10}}} & 81.06 $\pm$ 0.90\,\ensuremath{\uparrow_{\textcolor{blue}{\scriptsize 7.74}}} & \textbf{-5.81 $\pm$ 2.54}\,\ensuremath{\uparrow_{\textcolor{blue}{\scriptsize 11.49}}} & 59.66 $\pm$ 2.71\,\ensuremath{\uparrow_{\textcolor{blue}{\scriptsize 0.37}}} & 59.37 $\pm$ 1.30\,\ensuremath{\uparrow_{\textcolor{blue}{\scriptsize 0.01}}} & 1.89 $\pm$ 4.28\,\ensuremath{\uparrow_{\textcolor{blue}{\scriptsize 2.85}}} \\
\hline
iCaRL~\cite{rebuffi2017icarl} & 67.59 $\pm$ 1.36 & 71.89 $\pm$ 1.93 & -16.57 $\pm$ 1.92 & 78.98 $\pm$ 1.28 & 79.28 $\pm$ 0.30 & -8.07 $\pm$ 1.33 & 54.28 $\pm$ 1.52 & 56.52 $\pm$ 0.32 & 1.54 $\pm$ 1.41 \\
+STAIL(Ours)  & 71.47 $\pm$ 2.51\,\ensuremath{\uparrow_{\textcolor{blue}{\scriptsize 3.88}}} & 75.21 $\pm$ 0.44\,\ensuremath{\uparrow_{\textcolor{blue}{\scriptsize 3.32}}} & -8.70 $\pm$ 1.74\,\ensuremath{\uparrow_{\textcolor{blue}{\scriptsize 7.87}}} & 80.23 $\pm$ 1.94\,\ensuremath{\uparrow_{\textcolor{blue}{\scriptsize 1.25}}} & \textbf{82.58 $\pm$ 1.15}\,\ensuremath{\uparrow_{\textcolor{blue}{\scriptsize 3.30}}} & -8.29 $\pm$ 2.20\,\ensuremath{\downarrow_{\scriptsize 0.22}} & 59.97 $\pm$ 1.82\,\ensuremath{\uparrow_{\textcolor{blue}{\scriptsize 5.69}}} & 58.32 $\pm$ 1.30\,\ensuremath{\uparrow_{\textcolor{blue}{\scriptsize 1.80}}} & 2.03 $\pm$ 2.82\,\ensuremath{\uparrow_{\textcolor{blue}{\scriptsize 0.49}}} \\
\hline
WA~\cite{zhao2020maintaining} & 67.23 $\pm$ 1.67 & 72.56 $\pm$ 0.56 & -18.46 $\pm$ 3.03 & 75.91 $\pm$ 0.48 & 78.26 $\pm$ 0.28 & -12.30 $\pm$ 0.77 & 59.11 $\pm$ 3.33 & 59.36 $\pm$ 1.74 & 0.19 $\pm$ 1.35 \\
+STAIL(Ours)  & 71.62 $\pm$ 2.21\,\ensuremath{\uparrow_{\textcolor{blue}{\scriptsize 4.39}}} & \textbf{75.28 $\pm$ 0.99}\,\ensuremath{\uparrow_{\textcolor{blue}{\scriptsize 2.72}}} & -6.46 $\pm$ 4.56\,\ensuremath{\uparrow_{\textcolor{blue}{\scriptsize 12.00}}} & 78.96 $\pm$ 3.59\,\ensuremath{\uparrow_{\textcolor{blue}{\scriptsize 3.05}}} & 81.20 $\pm$ 1.82\,\ensuremath{\uparrow_{\textcolor{blue}{\scriptsize 2.94}}} & -7.53 $\pm$ 4.08\,\ensuremath{\uparrow_{\textcolor{blue}{\scriptsize 4.77}}} & \textbf{61.29 $\pm$ 2.73}\,\ensuremath{\uparrow_{\textcolor{blue}{\scriptsize 2.18}}} & \textbf{60.54 $\pm$ 2.43}\,\ensuremath{\uparrow_{\textcolor{blue}{\scriptsize 1.18}}} & 0.49 $\pm$ 2.72\,\ensuremath{\uparrow_{\textcolor{blue}{\scriptsize 0.30}}} \\
\hline
TagFex~\cite{zheng2025task} & 72.19 $\pm$ 2.83 & 72.93 $\pm$ 1.52 & -4.90 $\pm$ 2.15 & 79.90 $\pm$ 0.06 & 79.36 $\pm$ 0.57 & -6.97 $\pm$ 0.75 & 55.17 $\pm$ 3.35 & 57.82 $\pm$ 1.93 & -1.42 $\pm$ 1.53 \\
+STAIL(Ours)  & \textbf{72.20 $\pm$ 1.08}\,\ensuremath{\uparrow_{\textcolor{blue}{\scriptsize 0.01}}} & 73.04 $\pm$ 1.80\,\ensuremath{\uparrow_{\textcolor{blue}{\scriptsize 0.11}}} & \textbf{-4.82 $\pm$ 0.66}\,\ensuremath{\uparrow_{\textcolor{blue}{\scriptsize 0.08}}} & \textbf{81.50 $\pm$ 0.30}\,\ensuremath{\uparrow_{\textcolor{blue}{\scriptsize 1.60}}} & 80.39 $\pm$ 0.48\,\ensuremath{\uparrow_{\textcolor{blue}{\scriptsize 1.03}}} & -6.75 $\pm$ 0.89\,\ensuremath{\uparrow_{\textcolor{blue}{\scriptsize 0.22}}} & 58.83 $\pm$ 4.64\,\ensuremath{\uparrow_{\textcolor{blue}{\scriptsize 3.66}}} & 59.79 $\pm$ 2.41\,\ensuremath{\uparrow_{\textcolor{blue}{\scriptsize 1.97}}} & -2.36 $\pm$ 0.78\,\ensuremath{\downarrow_{\scriptsize 0.94}} \\

\hline
\end{tabular}%
}

\end{table*}

\begin{table*}[t]
\centering
\caption{Ablation Study of LSAM Components on Three Medical Datasets.}
\label{tab:ablation}
\renewcommand{\arraystretch}{1.3}

\setlength{\tabcolsep}{3pt} 
\resizebox{\textwidth}{!}{%
\begin{tabular}{l|ccc|ccc|ccc|ccc}
\hline
\multirow{2}{*}{\textbf{Method}} & \multicolumn{3}{c|}{\textbf{Components}} & \multicolumn{3}{c|}{\textbf{ODIR-5K}} & \multicolumn{3}{c|}{\textbf{US-DATA}} & \multicolumn{3}{c}{\textbf{MS-CXR}} \\
\cline{2-13}
 & $\mathcal{L}_{DSA}$ & $\mathcal{L}_{EPA}$ & $\mathcal{L}_{CSE}$ & Avg-AUC & AAA-AUC & BWT-AUC & Avg-AUC & AAA-AUC & BWT-AUC & Avg-AUC & AAA-AUC & BWT-AUC \\
\hline
\multirow{4}{*}{Replay} 
 & - & - & - & 66.97 $\pm$ 3.67 & 71.81 $\pm$ 1.12 & -17.65 $\pm$ 6.35 & 70.64 $\pm$ 0.75 & 73.32 $\pm$ 0.18 & -17.30 $\pm$ 1.37 & 59.29 $\pm$ 3.01 & 59.36 $\pm$ 1.69 & -0.96 $\pm$ 2.64 \\ 
 & \checkmark & - & - & 62.66 $\pm$ 5.45 & 71.25 $\pm$ 2.64 & -22.19 $\pm$ 4.97 & \textbf{85.09 $\pm$ 0.65} & \textbf{83.27 $\pm$ 0.42} & \textbf{-3.58 $\pm$ 0.76} & 57.15 $\pm$ 1.98 & 58.57 $\pm$ 1.06 & \textbf{2.31 $\pm$ 5.30} \\ 
 & \checkmark & \checkmark & - & 59.49 $\pm$ 1.34 & 70.09 $\pm$ 0.84 & -26.85 $\pm$ 1.14 & 83.74 $\pm$ 0.40 & 83.01 $\pm$ 0.54 & -4.50 $\pm$ 0.59 & 59.11 $\pm$ 3.50 & \textbf{59.56 $\pm$ 1.79} & -0.93 $\pm$ 3.99 \\ 
 & \checkmark & \checkmark & \checkmark & \textbf{67.60 $\pm$ 1.44} & \textbf{72.53 $\pm$ 1.03} & \textbf{-14.01 $\pm$ 2.80} & 78.74 $\pm$ 0.48 & 81.06 $\pm$ 0.90 & -5.81 $\pm$ 2.54 & \textbf{59.66 $\pm$ 2.71} & 59.37 $\pm$ 1.30 & 1.89 $\pm$ 4.28 \\ 
\hline
\multirow{4}{*}{iCaRL} 
 & - & - & - & 67.59 $\pm$ 1.36 & 71.89 $\pm$ 1.93 & -16.57 $\pm$ 1.92 & 78.98 $\pm$ 1.28 & 79.28 $\pm$ 0.30 & -8.07 $\pm$ 1.33 & 54.28 $\pm$ 1.52 & 56.52 $\pm$ 0.32 & 1.54 $\pm$ 1.41 \\
 & \checkmark & - & - & 68.44 $\pm$ 0.28 & 73.98 $\pm$ 0.81 & -11.95 $\pm$ 2.09 & 83.49 $\pm$ 0.20 & \textbf{84.29 $\pm$ 0.20} & -5.95 $\pm$ 0.40 & 57.32 $\pm$ 1.63 & 57.67 $\pm$ 2.46 & -0.25 $\pm$ 1.37 \\
 & \checkmark & \checkmark & - & 67.48 $\pm$ 1.81 & 73.81 $\pm$ 0.67 & -17.34 $\pm$ 2.60 & \textbf{84.01 $\pm$ 0.52} & 84.14 $\pm$ 0.29 & \textbf{-5.23 $\pm$ 1.07} & 55.15 $\pm$ 2.78 & 56.11 $\pm$ 2.99 & 1.23 $\pm$ 1.43 \\
 & \checkmark & \checkmark & \checkmark & \textbf{71.47 $\pm$ 2.51} & \textbf{75.21 $\pm$ 0.44} & \textbf{-8.70 $\pm$ 1.74} & 80.23 $\pm$ 1.94 & 82.58 $\pm$ 1.15 & -8.29 $\pm$ 2.20 & \textbf{59.97 $\pm$ 1.82} & \textbf{58.32 $\pm$ 1.30} & \textbf{2.03 $\pm$ 2.82} \\
\hline
\multirow{4}{*}{WA} 
 & - & - & - & 67.23 $\pm$ 1.67 & 72.56 $\pm$ 0.56 & -18.46 $\pm$ 3.03 & 75.91 $\pm$ 0.48 & 78.26 $\pm$ 0.28 & -12.30 $\pm$ 0.77 & 59.11 $\pm$ 3.33 & 59.36 $\pm$ 1.74 & 0.19 $\pm$ 1.35 \\
 & \checkmark & - & - & 67.95 $\pm$ 2.66 & 73.88 $\pm$ 1.18 & -12.41 $\pm$ 3.69 & 81.03 $\pm$ 0.89 & 82.70 $\pm$ 0.05 & -8.47 $\pm$ 1.10 & 58.47 $\pm$ 1.91 & 60.23 $\pm$ 1.54 & -0.20 $\pm$ 7.19 \\
 & \checkmark & \checkmark & - & 67.66 $\pm$ 2.12 & 73.89 $\pm$ 0.53 & -14.46 $\pm$ 1.84 & \textbf{82.32 $\pm$ 0.37} & \textbf{82.91 $\pm$ 0.18} & \textbf{-6.69 $\pm$ 0.42} & 58.25 $\pm$ 1.47 & 59.47 $\pm$ 1.87 & \textbf{0.98 $\pm$ 1.58} \\
 & \checkmark & \checkmark & \checkmark & \textbf{71.62 $\pm$ 2.21} & \textbf{75.28 $\pm$ 0.99} & \textbf{-6.46 $\pm$ 4.56} & 78.96 $\pm$ 3.59 & 81.20 $\pm$ 1.82 & -7.53 $\pm$ 4.08 & \textbf{61.29 $\pm$ 2.73} & \textbf{60.54 $\pm$ 2.43} & 0.49 $\pm$ 2.72 \\
\hline
\multirow{4}{*}{TagFex} 
 & - & - & - & 72.19 $\pm$ 2.83 & 72.93 $\pm$ 1.52 & -4.90 $\pm$ 2.15 & 79.90 $\pm$ 0.06 & 79.36 $\pm$ 0.57 & -6.97 $\pm$ 0.75 & 55.17 $\pm$ 3.35 & 57.82 $\pm$ 1.93 & \textbf{-1.42 $\pm$ 1.53} \\
 & \checkmark & - & - & 70.73 $\pm$ 1.56 & \textbf{73.22 $\pm$ 1.78} & -7.57 $\pm$ 0.82 & \textbf{81.67 $\pm$ 0.95} & \textbf{80.64 $\pm$ 0.48} & -6.93 $\pm$ 0.62 & 55.22 $\pm$ 0.94 & 58.59 $\pm$ 0.18 & -1.88 $\pm$ 1.02 \\
 & \checkmark & \checkmark & - & 71.48 $\pm$ 1.30 & 72.61 $\pm$ 2.43 & -5.96 $\pm$ 2.01 & 81.27 $\pm$ 0.46 & 80.39 $\pm$ 0.05 & -6.99 $\pm$ 0.67 & 55.20 $\pm$ 6.64 & 58.71 $\pm$ 2.36 & -1.66 $\pm$ 0.50 \\
 & \checkmark & \checkmark & \checkmark & \textbf{72.20 $\pm$ 1.08} & 73.04 $\pm$ 1.80 & \textbf{-4.82 $\pm$ 0.66} & 81.50 $\pm$ 0.30 & 80.39 $\pm$ 0.48 & \textbf{-6.75 $\pm$ 0.89} & \textbf{58.83 $\pm$ 4.64} & \textbf{59.79 $\pm$ 2.41} & -2.36 $\pm$ 0.78 \\
\hline
\end{tabular}%
}
\end{table*}

\subsubsection{Metrics and Baseline Methods}
Given the prevalence of class imbalance in medical datasets, we employ the Area Under the ROC Curve (AUC) as the primary metric. Let $T$ denote the total number of sequential tasks. We define $A_{i,j}$ as the test AUC on task $j$ after the model has finished training on task $i$ (where $j \leq i$). We utilize three metrics~\cite{wu2024meta} to comprehensively evaluate the incremental learning performance:

\textbf{Average AUC (Avg-AUC)} evaluates the final global performance across all tasks after the entire training sequence is completed.
\begin{equation}
    \text{Avg-AUC} = \frac{1}{T} \sum_{j=1}^{T} A_{T, j}.
\end{equation}

\textbf{Average Anytime AUC (AAA-AUC)~\cite{caccia2021new}} assesses the model's performance consistency throughout the continuous learning stream. A higher AAA-AUC indicates robust performance during the entire learning process, not just at the end.
\begin{equation}
    \text{AAA-AUC} = \frac{1}{T} \sum_{i=1}^{T} \left( \frac{1}{i} \sum_{j=1}^{i} A_{i, j} \right).
\end{equation}

\textbf{Backward Transfer AUC (BWT-AUC)\cite{11329432}} quantifies the impact of learning new tasks on the forgetting of old knowledge.
\begin{equation}
    \text{BWT-AUC} = \frac{1}{T-1} \sum_{i=1}^{T-1} (A_{T, i} - A_{i, i}).
\end{equation}

To comprehensively validate our proposed framework, we compare it against several CIL baselines spanning different technical trajectories: regularization-based (EWC~\cite{kirkpatrick2017overcoming}), architecture-based (DER~\cite{yan2021dynamically} and MEMO~\cite{zhoumodel}), rehearsal-based (Replay, iCaRL~\cite{rebuffi2017icarl}, WA~\cite{zhao2020maintaining} , and TagFex~\cite{zheng2025task}), and Prompt-based methods (L2P~\cite{wang2022L2P} and DualPrompt~\cite{wang2022dualprompt}).

To demonstrate the versatility and effectiveness of our approach, we explicitly integrate our STAIL framework as a plug-and-play module into the aforementioned rehearsal-based methods for direct comparative analysis.

\subsubsection{Implementation details}
For prompt-based methods, we followed their original frozen-backbone training protocols. All conventional CIL baselines were implemented based on the PyCIL~\cite{zhou2023pycil}. For the visual stream, we uniformly employed ResNet-18~\cite{he2016deep} as the backbone network. For the textual stream, we utilized the biomedical-specific LLM, BioMistral-7B~\cite{labrak2024biomistral}, which remained frozen throughout the entire training process. The model's trainable parameters were optimized using the Adam optimizer. During training, the image batch size is set to 64. For the contrastive objectives $\mathcal{L}_{EPA}$ and $\mathcal{L}_{CSE}$, the image batch size $B_I$ and text batch size $B_S$ are set to 64 and 128, respectively. We maintain a fixed per-class memory budget to ensure fair comparisons across all competing methods. To ensure reproducibility, we repeated all experiments across 3 different random seeds and reported the results as mean $\pm$ standard deviation. We provided more implementation details in Appendix~\ref{Supplementary Implementation Details}.
\subsection{Experimental Results}
\textbf{Quantitative Results}. As shown in Table~\ref{table:main_results}, the STAIL framework consistently outperforms mainstream incremental learning methods across three diverse medical datasets, demonstrating its comprehensive effectiveness in enhancing overall accuracy, maintaining stability, and suppressing forgetting. On the ODIR-5K fundus dataset, STAIL propelled the WA baseline to establish a new SOTA in learning stability (AAA-AUC 75.28\%) while simultaneously driving a substantial 12.00\% increase in BWT-AUC. Furthermore, iCaRL+STAIL achieved breakthroughs through effective semantic anchoring. This robustness extends to the multi-organ US-DATA ultrasound dataset, where STAIL assisted the TagFex method in achieving highly competitive performance (Avg-AUC 81.50\%). It also helped Replay navigate 41 fine-grained categories and cross-organ domain shifts by leveraging textual semantics to mitigate image noise and inter-class similarity. On the MS-CXR dataset, our approach fully exploited image-text alignment to reach superior results, notably achieving Positive Backward Transfer (BWT-AUC 2.03\% for iCaRL+STAIL). This rare phenomenon is primarily attributed to the review effect via buffer dominance~\cite{lin2022beyond,liturning} in long-tail distributions, where the model effectively consolidates old knowledge during later incremental stages. 

Beyond conventional CIL approaches, we further compare STAIL with recent prompt-based continual learning methods, including L2P and DualPrompt, which adapt pretrained vision-language models by freezing the visual backbone and optimizing learnable prompts. STAIL consistently outperforms these prompt-based competitors across all three medical benchmarks. For example, on the US-DATA dataset, TagFex+STAIL achieves an Avg-AUC of 81.50\%, exceeding L2P and DualPrompt by 4.40\% and 5.01\%, respectively. These improvements demonstrate that image-level clinical text anchors provide richer semantic guidance than class-level prompts, enabling more effective preservation and adaptation of fine-grained medical knowledge during incremental learning. Although prompt-based methods benefit from strong pretrained representations, their frozen visual encoders may limit adaptation to evolving medical distributions. In contrast, STAIL jointly optimizes the visual representation and leverages stable LLM-derived semantic constraints, achieving a better balance between plasticity and stability. Collectively, these results confirm that STAIL provides a robust and highly competitive solution for medical multimodal incremental learning.

\textbf{Ablation Study}. The ablation study presented in Table~\ref{tab:ablation} investigates the individual contributions of the three components within the LLM-derived Semantic Anchoring Mechanism (LSAM): $\mathcal{L}_{DSA}$, $\mathcal{L}_{EPA}$, and $\mathcal{L}_{CSE}$. The results demonstrate that these components provide complementary functions for balancing representation stability and plasticity during continual medical learning. The results indicate that $\mathcal{L}_{DSA}$ serves as a critical foundation by aligning evolving visual features with global LLM semantic priors to prevent macroscopic drift. This is particularly evident on the high-noise US-DATA dataset, where its integration into the Replay baseline increased the Avg-AUC from 70.64\% to 85.09\%. The superior performance of $\mathcal{L}_{DSA}$ alone on this dataset can be attributed to the challenging characteristics of US-DATA, which contains 41 fine-grained categories across multiple anatomical organs, where strong semantic regularization provides effective guidance for final representation optimization. However, due to the heterogeneous and fine-grained characteristics of multi-organ medical data, where different anatomical domains exhibit distinct visual patterns and clinical variations, the complete LSAM framework does not consistently outperform individual variants on every metric. For such complex clinical scenarios, future work could investigate organ-aware hierarchical semantic representations, where organ-level anatomical priors~\cite{shui2025large} are integrated with disease-specific pathological patterns to establish more transferable medical knowledge and improve knowledge retention in multi-organ continual learning scenarios.

Building upon these macroscopic anchors, $\mathcal{L}_{EPA}$ further refines feature discriminability to ensure the model's plasticity for new knowledge, as evidenced by the WA baseline's Avg-AUC elevation to 82.32\% on US-DATA. While $\mathcal{L}_{EPA}$ improves the model's ability to acquire discriminative representations for emerging categories, maintaining previously learned knowledge remains challenging during incremental updates. Therefore, $\mathcal{L}_{CSE}$ further consolidates historical representations by leveraging semantic exemplars from previous tasks to preserve the structural integrity of old classes. This stabilizing effect is most prominent on the ODIR-5K dataset, where it improves BWT-AUC across baselines (e.g., improving the WA method from -14.46\% to -6.46\%). Ultimately, these results confirm that the three components within LSAM are complementary, collectively enabling the model to balance the acquisition of new clinical knowledge while preserving historical representations.

\begin{figure}[width=\linewidth,pos=t]
\centering

\begin{minipage}{\linewidth}
    \centering
    
    \includegraphics[width=0.48\linewidth]{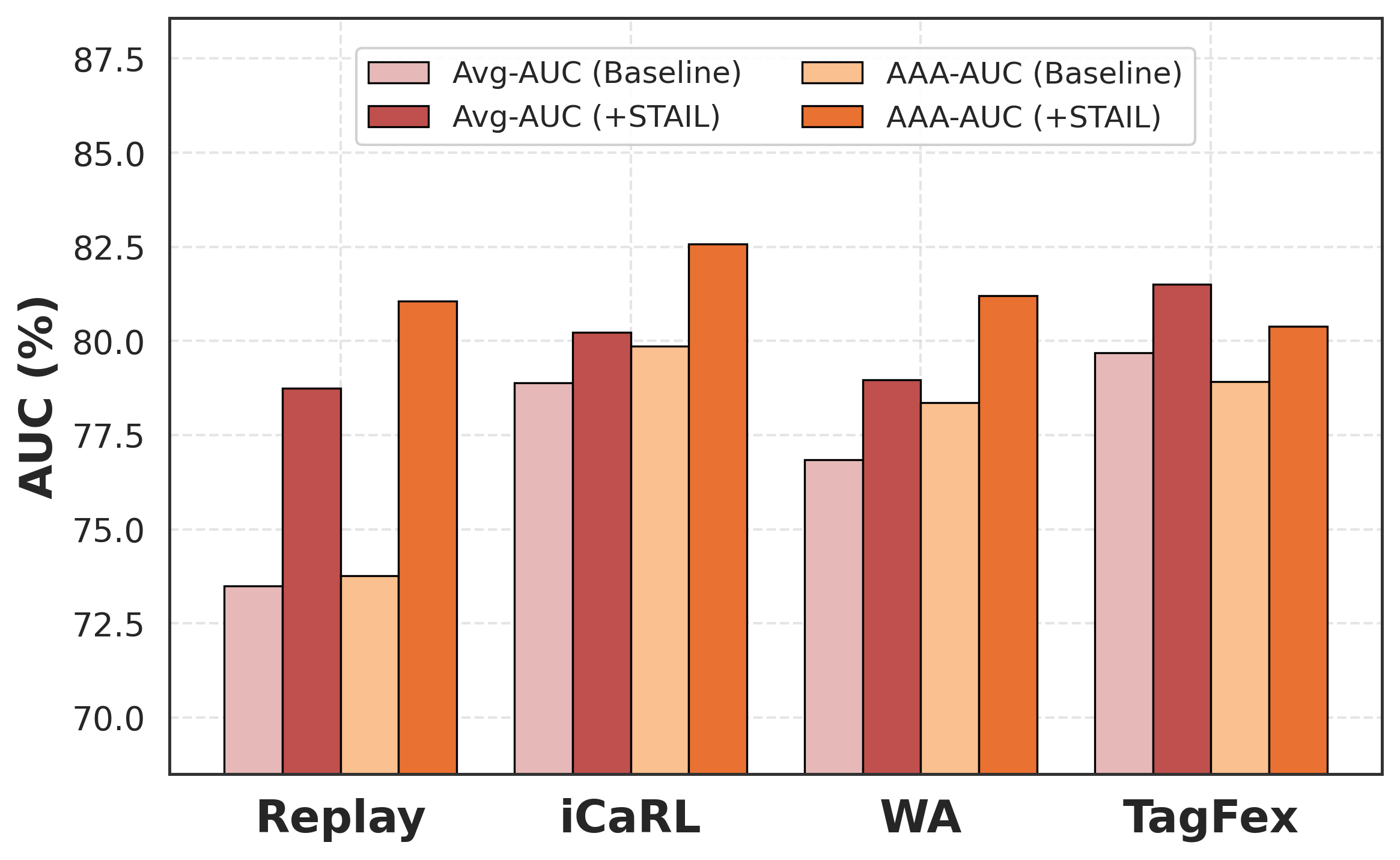}
    \hfill
    \includegraphics[width=0.48\linewidth]{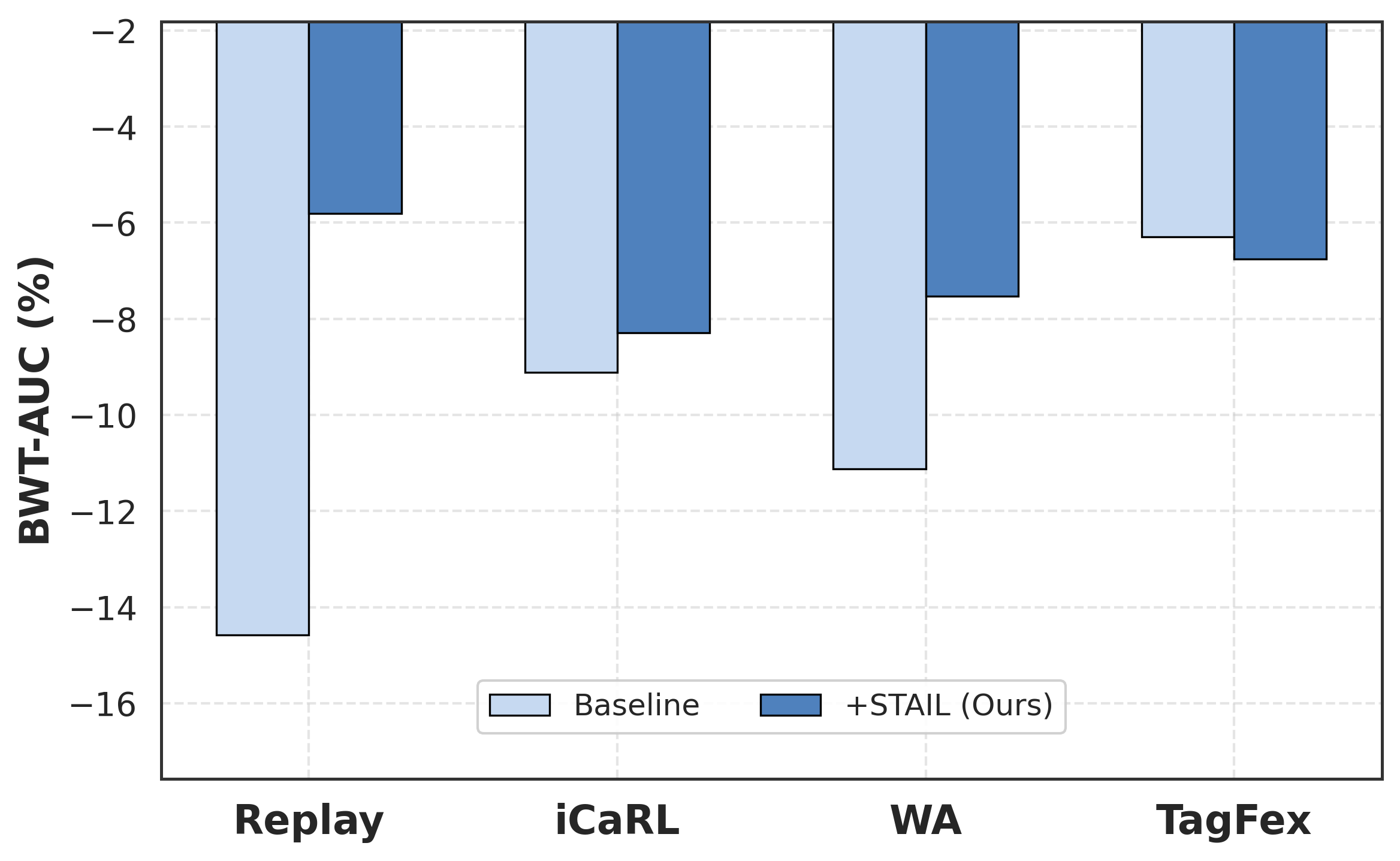}
    
    \vspace{0.5em}
    
    (a) US-DATA
\end{minipage}

\vspace{1.5em}

\begin{minipage}{\linewidth}
    \centering
    
    \includegraphics[width=0.48\linewidth]{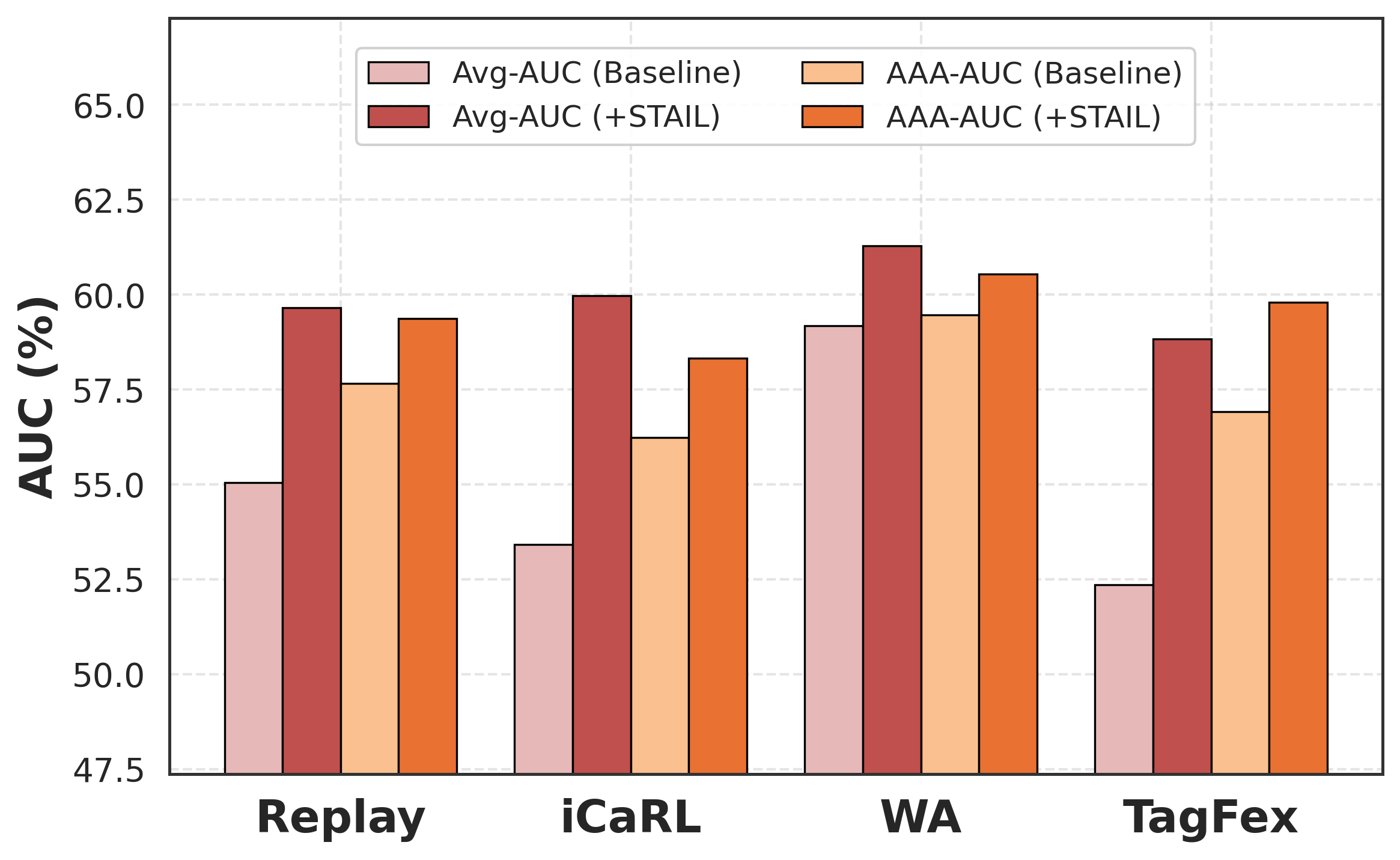}
    \hfill
    \includegraphics[width=0.48\linewidth]{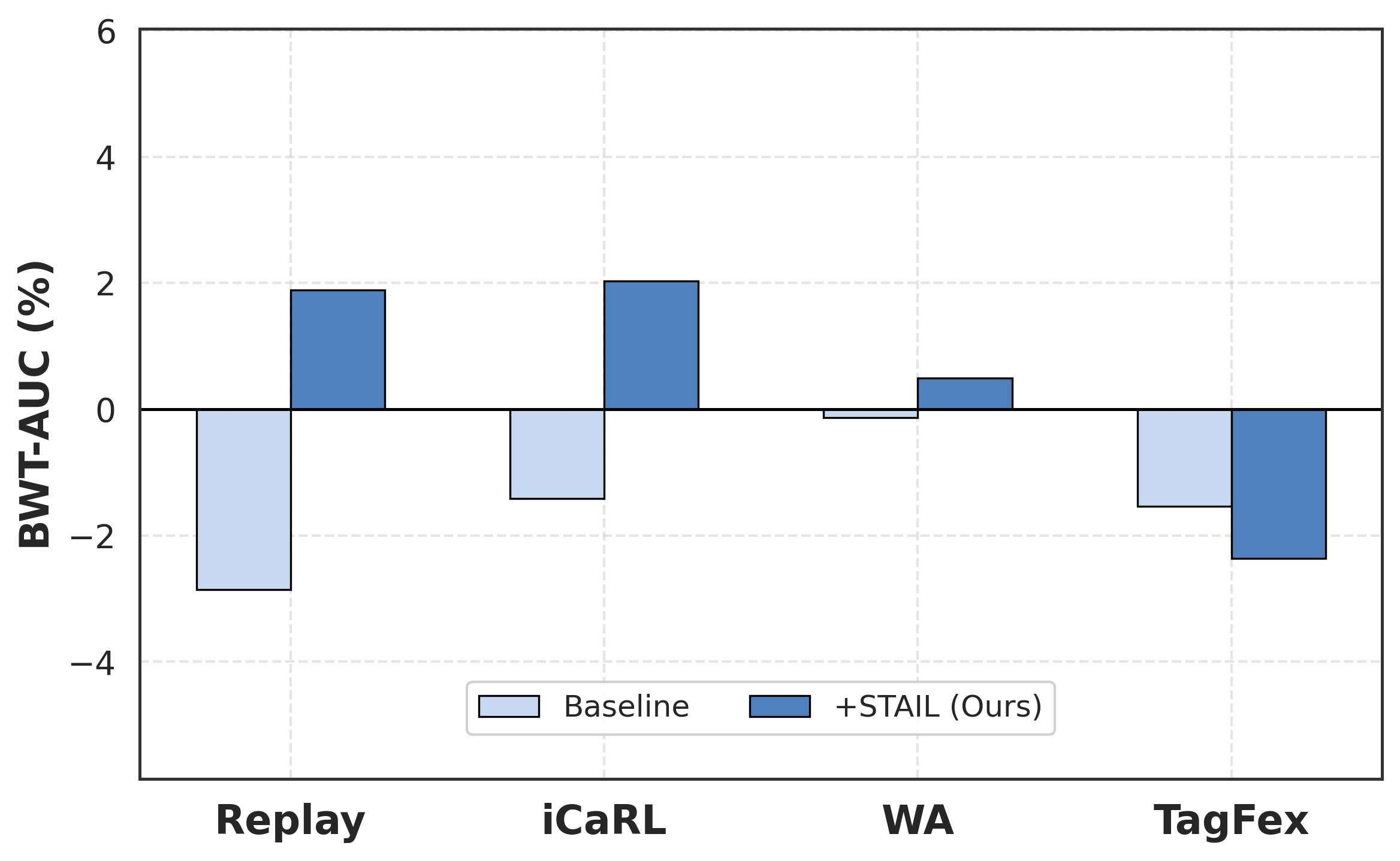}
    
    \vspace{0.5em}
    
    (b) MS-CXR
\end{minipage}

\caption{Performance comparison between baseline methods and their STAIL variants under equal buffer memory usage. Results are averaged over three runs.}
\label{fig:memory_comparison}

\end{figure}

\textbf{Experiments Under Equal Buffer Memory Usage}. Although practical implementations often rely on file paths, we adhere to standard incremental learning protocols~\cite{rebuffi2017icarl} by calculating budgets based on the theoretical byte size of raw media. Specifically, a single standard $224 \times 224$ RGB image consumes approximately 147 KB ($150,528$ Bytes). In stark contrast, our statistical analysis reveals that the average text description lengths for the US-DATA and MS-CXR datasets are merely 325.39 Bytes and 44.68 Bytes, respectively. This means that a single image requires approximately 462 times and 3,369 times more memory than a single text description in these respective datasets. This magnitude of difference implies that  the memory overhead introduced by our SCB strategy is negligible. To rigorously validate the practical efficacy of the STAIL framework under strict storage constraints, we designed a comparative experiment that slightly favors the baseline methods.

\begin{itemize}
\item \textbf{US-DATA}: The baseline stores 6 images per class, whereas STAIL stores only 5 images per class + 25 texts per class.
\item \textbf{MS-CXR}: The baseline stores 21 images per class, whereas STAIL stores only 20 images per class + 100 texts per class.
\end{itemize}

Consequently, the single extra image stored by the baseline consumes significantly more memory than all the text exemplars added by STAIL. In other words, the baseline method benefits from both a larger total memory budget and a greater number of visual training samples. However, even under the dual disadvantages of lower memory usage and fewer visual exemplars, the STAIL framework achieved substantial performance superiority. As shown in Fig. \ref{fig:memory_comparison}, STAIL achieves substantial gains in both anti-forgetting capabilities and classification accuracy, representing a highly cost-effective solution for medical incremental learning.

\textbf{Experiments Using Different Large Language Models}.
\begin{figure*}[t]
    \centering
    \includegraphics[width=0.3\linewidth]{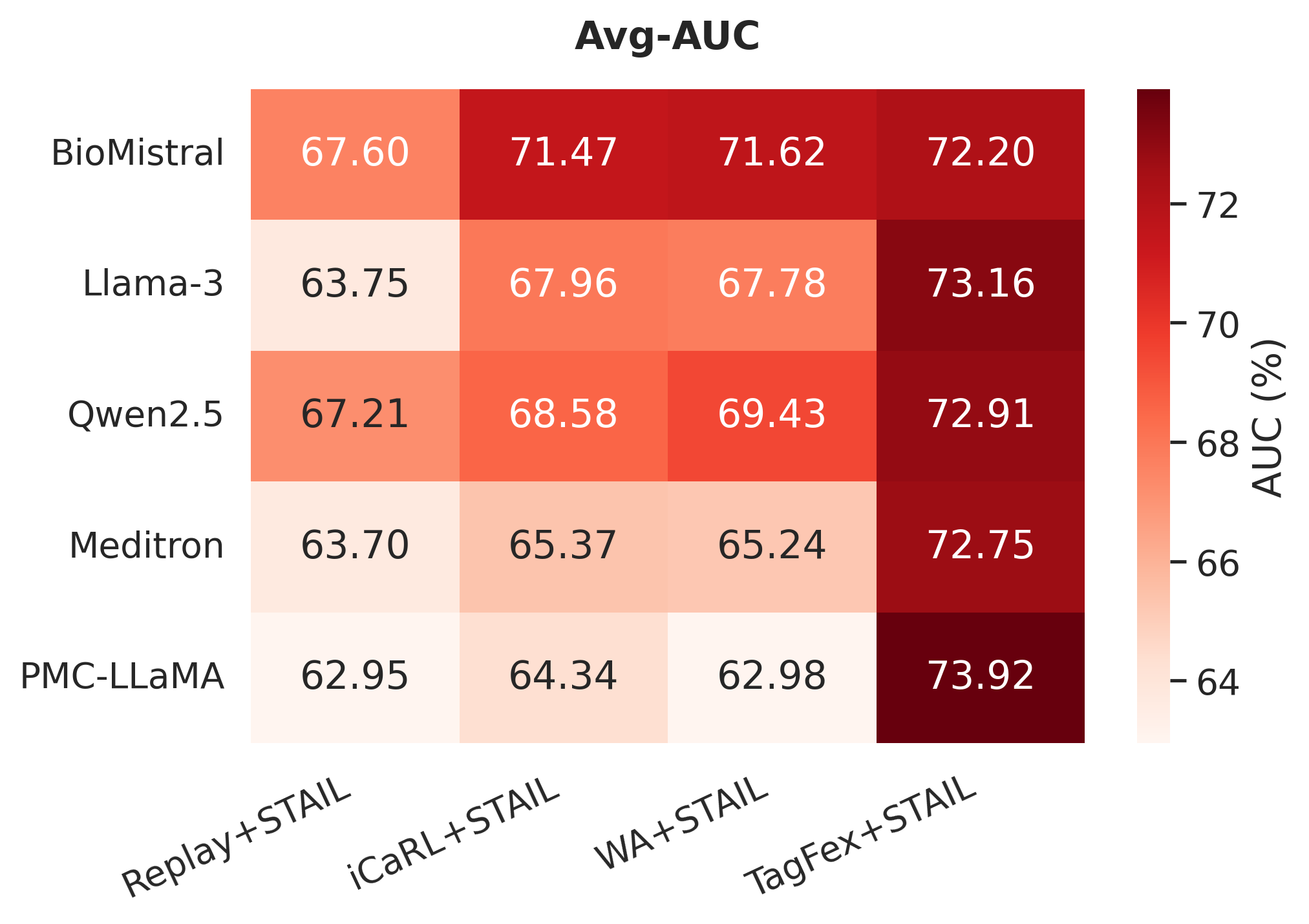}
    \hfill
    \includegraphics[width=0.3\linewidth]{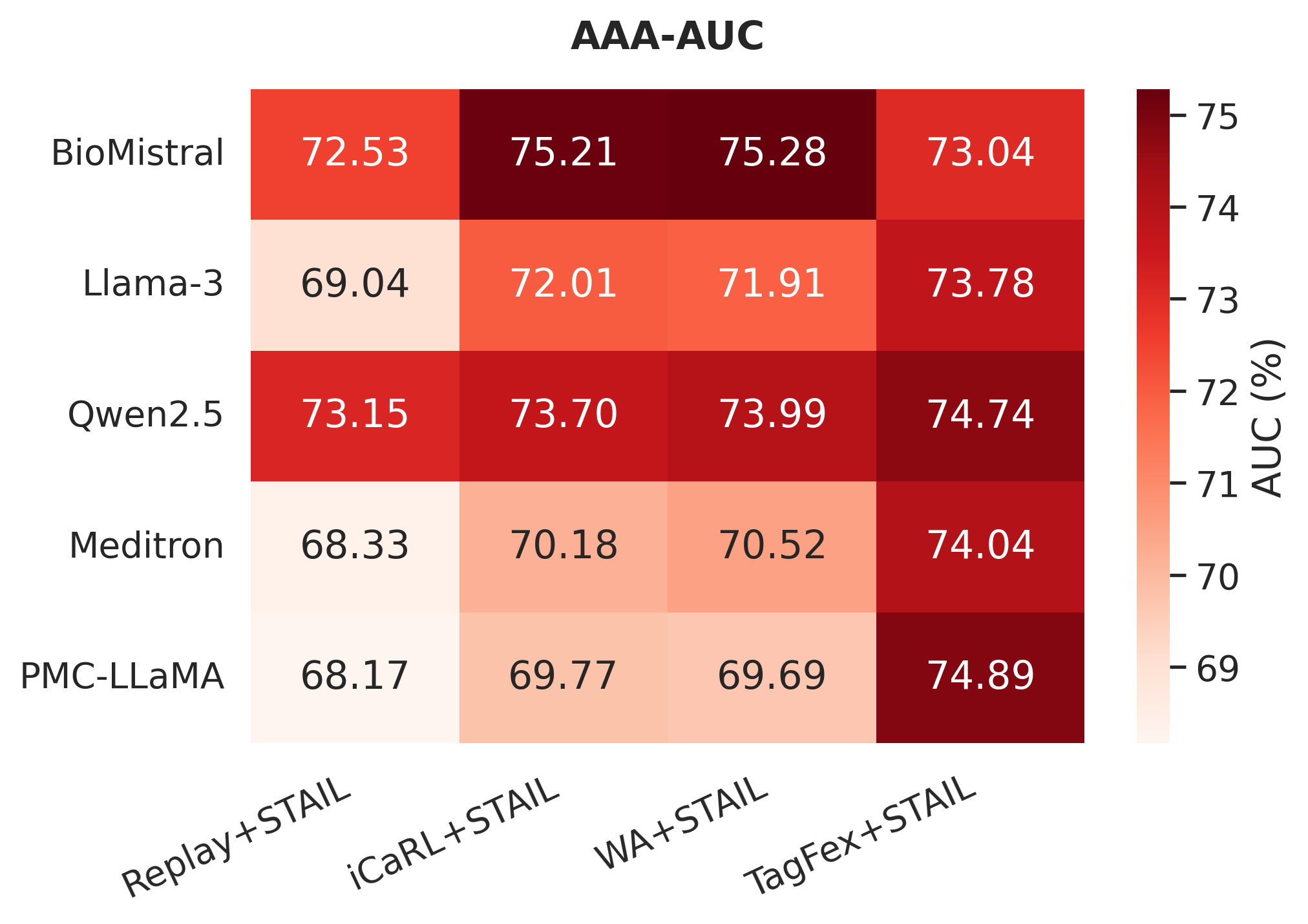}
    \hfill
    \includegraphics[width=0.3\linewidth]{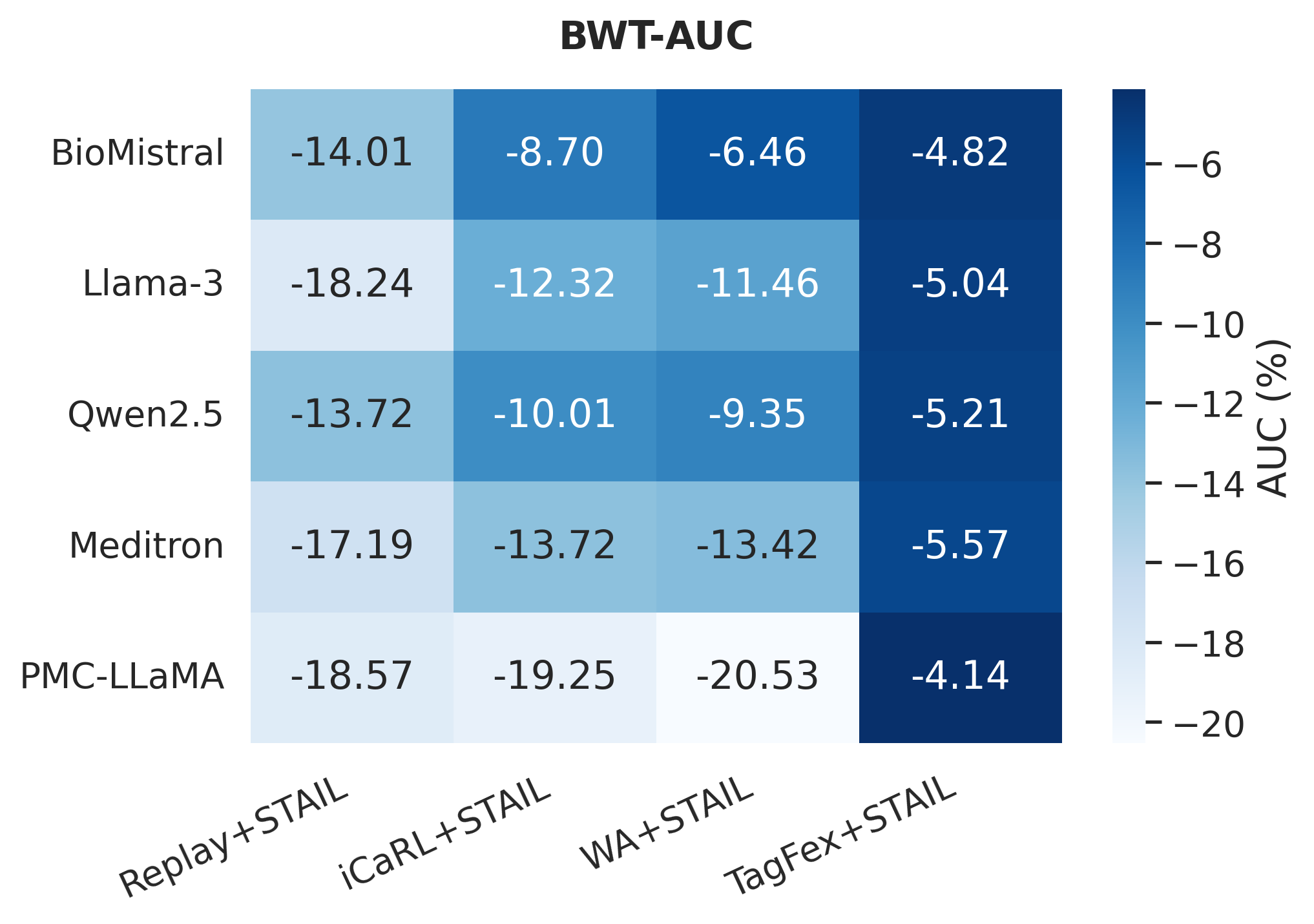}
    \caption{Comparison of different LLMs on the ODIR-5K dataset, averaged across three runs.}
    \label{fig:LLM_heatmaps}
\end{figure*}
To investigate the impact of different types of LLMs as semantic feature extractors on the performance of the STAIL framework, we conducted an extensive ablation study on the ODIR-5K dataset. The selected text models were categorized into general-domain SOTA models (Llama-3\cite{grattafiori2024llama}, Qwen2.5\cite{qwen2.5}) and medical-specific models (BioMistral\cite{labrak2024biomistral}, Meditron\cite{chen2023meditron}, PMC-LLaMA\cite{wu2024pmc}). As shown in Fig. \ref{fig:LLM_heatmaps}, the semantic priors provided by different LLMs exhibit significantly different feature-guidance effects when integrated with different baseline methods.

(1) Performance and Limitations of General-Domain LLMs.
While Llama-3 and Qwen2.5 showed robust performance with TagFex, with Qwen2.5 even surpassing some medical models on the Replay baseline (73.15\% AAA-AUC), they lagged behind BioMistral on fixed-architecture baselines such as iCaRL and WA. This is likely because general LLMs introduce excessive non-medical generalized semantics. For limited-capacity baselines, this semantic noise hinders the formation of sharp feature anchors for fundus lesions, ultimately weakening semantic alignment.

(2) Performance and Effects of Medical-Specific LLMs.
In contrast, medical-specific LLMs leverage massive medical corpora to provide high-quality domain priors. BioMistral demonstrated the best generalizability and robustness, consistently delivering stable improvements across all methods. This confirms its extracted semantic features are highly precise and easily aligned by conventional visual networks.

The performance of PMC-LLaMA warrants special attention. Although it suffered a substantial performance degradation on traditional baselines (Replay, iCaRL, WA), it achieved the highest peak performance when combined with the dynamic TagFex+STAIL architecture. Specifically, its Avg-AUC reached 73.92\%, AAA-AUC climbed to 74.89\%, and the BWT-AUC metric improved to an impressive -4.14\%.

\textbf{Experiments on Different Text Selection Strategies}. To evaluate the impact of text selection strategies within the SCB, we compared Random, Herding, and Entropy sampling on the ODIR-5K dataset (Table~\ref{tab:text_strategy_comparison}).
\begin{table*}[h]
\centering
\caption{Comparison of Different Text Selection Strategies on the ODIR-5K Dataset}
\label{tab:text_strategy_comparison}
\renewcommand{\arraystretch}{1.3} 
\setlength{\tabcolsep}{3pt}       

\resizebox{\textwidth}{!}{%
\begin{tabular}{l|ccc|ccc|ccc|ccc}
\hline
\multirow{2}{*}{\textbf{Text Strategy}} & \multicolumn{3}{c|}{\textbf{Replay+STAIL}} & \multicolumn{3}{c|}{\textbf{iCaRL+STAIL}} & \multicolumn{3}{c|}{\textbf{WA+STAIL}} & \multicolumn{3}{c}{\textbf{TagFex+STAIL}} \\
\cline{2-13} 
 & \textbf{Avg-AUC} & \textbf{AAA-AUC} & \textbf{BWT-AUC} & \textbf{Avg-AUC} & \textbf{AAA-AUC} & \textbf{BWT-AUC} & \textbf{Avg-AUC} & \textbf{AAA-AUC} & \textbf{BWT-AUC}  &\textbf{Avg-AUC} & \textbf{AAA-AUC} & \textbf{BWT-AUC} \\
\hline
\textbf{Random} & 67.60 $\pm$ 1.44 & 72.53 $\pm$ 1.03 & \textbf{-14.01 $\pm$ 2.80} & \textbf{71.47 $\pm$ 2.51} & \textbf{75.21 $\pm$ 0.44} & -8.70 $\pm$ 1.74 & \textbf{71.62 $\pm$ 2.21} & \textbf{75.28 $\pm$ 0.99} & \textbf{-6.46 $\pm$ 4.56} & 72.20 $\pm$ 1.08 & 73.04 $\pm$ 1.80 & -4.82 $\pm$ 0.66 \\
\hline
\textbf{Herding} & \textbf{67.85 $\pm$ 2.66} & \textbf{72.77 $\pm$ 0.46} & -14.40 $\pm$ 3.95 & 70.33 $\pm$ 3.02 & 74.59 $\pm$ 1.01 & \textbf{-6.53 $\pm$ 2.21} & 70.98 $\pm$ 2.48 & 74.86 $\pm$ 0.24 & -9.53 $\pm$ 5.08 & 69.98 $\pm$ 2.94 & 72.73 $\pm$ 2.41 & -5.38 $\pm$ 0.90 \\
\hline
\textbf{Entropy} &  67.05 $\pm$ 4.44 & 72.69 $\pm$ 1.46 & -16.79 $\pm$ 8.54 & 71.20 $\pm$ 1.45 & 74.80 $\pm$ 0.82 & -9.70 $\pm$ 3.13 & 70.91 $\pm$ 2.48 & 74.65 $\pm$ 0.63 & -9.71 $\pm$ 3.29 & \textbf{73.09 $\pm$ 1.30} & \textbf{74.83 $\pm$ 0.69} & \textbf{-3.87 $\pm$ 1.38} \\
\hline
\end{tabular}%
}
\end{table*}
\begin{figure*}[t]
    \centering
    \includegraphics[width=1.0\linewidth]{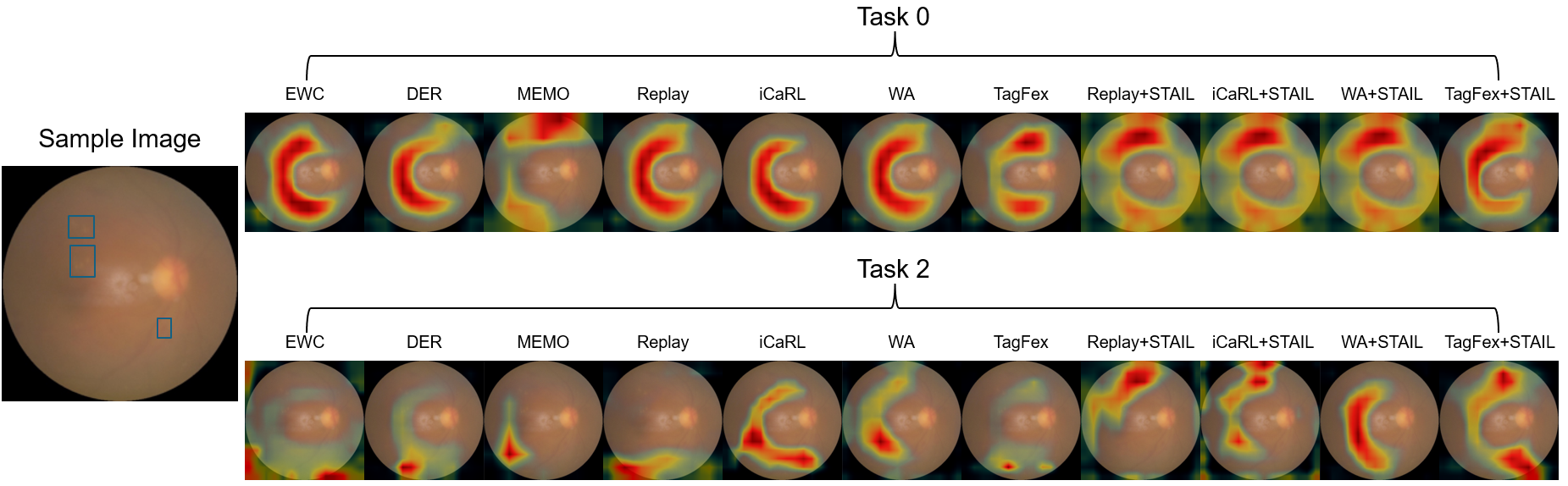}
    \caption{Activation maps on a cataract sample from the ODIR-5K dataset.}
    \label{fig:Activation maps_Cataract}
\end{figure*}

The experimental results indicate that Random sampling exhibits superior robustness and versatility across the different baselines. Specifically, it achieves optimal or highly competitive results in the majority of metrics, particularly for the WA+STAIL and iCaRL+STAIL settings. In specific scenarios, however, distinct behaviors were observed with other strategies. For instance, Herding yields a marginal performance gain (approximately 0.25\%) in both Avg-AUC and AAA-AUC when combined with Replay+STAIL. Additionally, the Entropy strategy allows TagFex+STAIL to reach peak performance, achieving 73.09\% Avg-AUC and a low -3.87\% BWT-AUC. Furthermore, a distinct advantage of Random sampling is its minimal computational overhead. Unlike Herding or Entropy, which require calculating distances or uncertainty scores, random selection is computationally negligible. Consequently, we adopt the Random strategy as the default setting for our framework to ensure efficiency and generalization capability.

\textbf{Visualization Results}. To explore how the model combats feature drift, we visualized feature activation maps for an ODIR-5K cataract sample using Grad-CAM~\cite{selvaraju2017grad}. Ophthalmologists manually annotated the core lesions (blue bounding boxes) to establish a clinical gold standard for evaluating the model's focus. As shown in Fig. \ref{fig:Activation maps_Cataract}, all models initially focus correctly on the lesion during Task 0. However, after training on Task 2, baseline methods suffer severe feature drift, erroneously shifting focus to the optic disc or image edges and failing to identify the cataract. In contrast, variants equipped with STAIL demonstrate robust performance. Their high-response regions remain accurately localized within the annotated boxes across all incremental updates. This clinical visualization highlights STAIL's interpretability. It demonstrates that by utilizing the SCB and LLM semantic anchoring, the framework effectively locks visual features onto the correct lesion manifold, successfully preventing feature space distortion typically induced by new tasks.

\subsection{Additional Experiment}
\label{Additional Experiment}
\begin{figure*}
    \centering
    \includegraphics[width=1\linewidth]{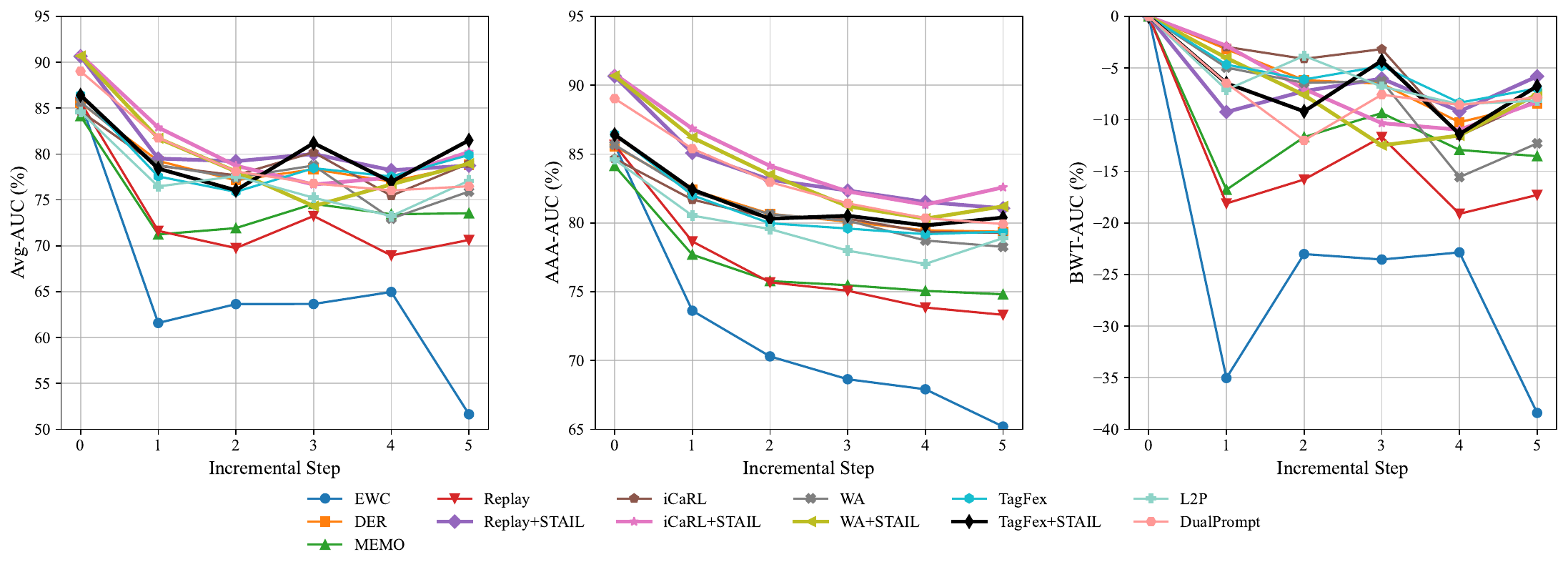}
    \caption{Avg-AUC, AAA-AUC, and BWT-AUC at each incremental step on US-DATA, averaged across three runs.}
    \label{fig:Incremental_Steps}
\end{figure*}
\textbf{Dynamic Performance Across Incremental Steps}. To intuitively illustrate the dynamic performance of the models during the incremental learning process, Fig. \ref{fig:Incremental_Steps} presents the trajectories of the evaluation metrics across incremental steps (Steps 0 to 5) on the US-DATA dataset. As depicted, the performance of traditional baseline methods inevitably deteriorates with the continuous influx of new tasks. Notably, non-rehearsal method EWC experiences a drastic performance collapse. However, methods equipped with the STAIL framework consistently maintain top-tier performance throughout the entire incremental learning trajectory. Particularly on the AAA-AUC curves, which reflect the global stability of the learning stream, the STAIL variants not only sustain the highest performance levels but also exhibit the smoothest degradation trends. Regarding BWT-AUC, baseline methods exhibit severe volatility and catastrophic forgetting when confronted with continuous distribution shifts, such as the original Replay dropping sharply below -15\%. In contrast, the integration of STAIL substantially mitigates this decline, effectively bounding the BWT-AUC within a minimal negative margin. This dynamic and intuitive visualization compellingly demonstrates the robust anti-forgetting capability of our semantic buffer throughout long-sequence learning.

\textbf{Different Category Sequential Experiment}.
\begin{table*}[h]
\centering
\caption{Comparison of Different Category Sequences on the MS-CXR Dataset}
\label{table: different category sequential}
\renewcommand{\arraystretch}{1.3} 
\setlength{\tabcolsep}{3pt}
\resizebox{\textwidth}{!}{%
\begin{tabular}{l|ccc|ccc|ccc}
\hline
\multirow{2}{*}{\textbf{Method}} & \multicolumn{3}{c|}{\textbf{Tail classes in Task 2}} & \multicolumn{3}{c|}{\textbf{Tail classes in Task 1}} & \multicolumn{3}{c}{\textbf{Tail classes in Task 0}} \\
\cline{2-10}
 & Avg-AUC & AAA-AUC & BWT-AUC & Avg-AUC & AAA-AUC & BWT-AUC & Avg-AUC & AAA-AUC & BWT-AUC \\ 
\hline
EWC~\cite{kirkpatrick2017overcoming}& 61.43  & 55.54 & \textbf{8.27}& 54.06 & 58.95 & 0.72  &  57.76 & 58.45 & -8.42  \\
\hline
DER~\cite{yan2021dynamically} & 55.73  & 57.41  & -1.57  & 56.21  & 62.68  & -4.13  & 60.38  & 63.14  & -2.80  \\
\hline
MEMO~\cite{zhoumodel} & 57.33  & 60.75  & -4.21  & 59.17  & 60.37  & -0.81  &  59.72 & \textbf{64.50} & -15.44  \\
\hline
L2P~\cite{wang2022L2P} & 56.56  & 58.83  & 0.02 & 60.34 & 60.85  & 0.64  &  63.00 & 63.60 & -1.29  \\
\hline
DualPrompt~\cite{wang2022dualprompt} & 56.09  & 57.38  & -0.17  & 59.65  & 62.37  & -0.75  &  62.49 & 63.36 & -1.82  \\
\hline
Replay  & 57.50  & 57.87 & -2.08  & 50.94  & 58.45  & -10.34  & 58.84  & 60.99  & -4.67  \\
+STAIL(Ours)  & 59.40\,\ensuremath{\uparrow_{\textcolor{blue}{\scriptsize 1.90}}} & 58.03\,\ensuremath{\uparrow_{\textcolor{blue}{\scriptsize 0.16}}} & 6.02\,\ensuremath{\uparrow_{\textcolor{blue}{\scriptsize 8.10}}} & 56.68\,\ensuremath{\uparrow_{\textcolor{blue}{\scriptsize 5.74}}} & 59.51\,\ensuremath{\uparrow_{\textcolor{blue}{\scriptsize 1.06}}} & -2.82\,\ensuremath{\uparrow_{\textcolor{blue}{\scriptsize 7.52}}} & 59.54\,\ensuremath{\uparrow_{\textcolor{blue}{\scriptsize 0.70}}} & 61.95\,\ensuremath{\uparrow_{\textcolor{blue}{\scriptsize 0.96}}} & -5.52\,\ensuremath{\downarrow_{\scriptsize 0.85}} \\
\hline
iCaRL~\cite{rebuffi2017icarl} & 55.99  & 56.18  & 2.49  & 53.39  & 58.69  & -0.29  & 55.91 & 59.05 & -8.01  \\
+STAIL(Ours)  & 62.06\,\ensuremath{\uparrow_{\textcolor{blue}{\scriptsize 6.07}}} & 58.92\,\ensuremath{\uparrow_{\textcolor{blue}{\scriptsize 2.74}}} & 3.79\,\ensuremath{\uparrow_{\textcolor{blue}{\scriptsize 1.30}}} & 53.84\,\ensuremath{\uparrow_{\textcolor{blue}{\scriptsize 0.45}}} & 59.48\,\ensuremath{\uparrow_{\textcolor{blue}{\scriptsize 0.79}}} & 0.02\,\ensuremath{\uparrow_{\textcolor{blue}{\scriptsize 0.31}}} & 59.10\,\ensuremath{\uparrow_{\textcolor{blue}{\scriptsize 3.19}}} & 62.02\,\ensuremath{\uparrow_{\textcolor{blue}{\scriptsize 2.97}}} & -5.96\,\ensuremath{\uparrow_{\textcolor{blue}{\scriptsize 2.05}}} \\
\hline
WA~\cite{zhao2020maintaining} & 58.74  & 58.72  & -1.02  & 58.28  & 61.01  & \textbf{1.25}  & 56.62  & 60.29  & -10.30  \\
+STAIL(Ours)  & 60.92\,\ensuremath{\uparrow_{\textcolor{blue}{\scriptsize 2.18}}} & 59.11\,\ensuremath{\uparrow_{\textcolor{blue}{\scriptsize 0.39}}} & 1.95\,\ensuremath{\uparrow_{\textcolor{blue}{\scriptsize 2.97}}} & \textbf{60.46}\,\ensuremath{\uparrow_{\textcolor{blue}{\scriptsize 2.18}}} & 62.28\,\ensuremath{\uparrow_{\textcolor{blue}{\scriptsize 1.27}}} & 0.98\,\ensuremath{\downarrow_{\scriptsize 0.27}} & \textbf{64.19}\,\ensuremath{\uparrow_{\textcolor{blue}{\scriptsize 7.57}}} & \textbf{64.48}\,\ensuremath{\uparrow_{\textcolor{blue}{\scriptsize 4.19}}} & \textbf{-1.12}\,\ensuremath{\uparrow_{\textcolor{blue}{\scriptsize 9.18}}} \\
\hline
TagFex~\cite{zheng2025task} & 51.53  & 56.38  & -2.15 & 52.82  & 58.03  & -1.76  & 58.87 & 60.55  & -6.01 \\
+STAIL(Ours)  & \textbf{63.50}\,\ensuremath{\uparrow_{\textcolor{blue}{\scriptsize 11.97}}} & \textbf{61.57}\,\ensuremath{\uparrow_{\textcolor{blue}{\scriptsize 5.19}}} & -1.96\,\ensuremath{\uparrow_{\textcolor{blue}{\scriptsize 0.19}}} & 59.48\,\ensuremath{\uparrow_{\textcolor{blue}{\scriptsize 6.66}}} & \textbf{63.17}\,\ensuremath{\uparrow_{\textcolor{blue}{\scriptsize 5.14}}} & -1.19\,\ensuremath{\uparrow_{\textcolor{blue}{\scriptsize 0.57}}} & 60.77\,\ensuremath{\uparrow_{\textcolor{blue}{\scriptsize 1.90}}} & 61.03\,\ensuremath{\uparrow_{\textcolor{blue}{\scriptsize 0.48}}} & -8.66\,\ensuremath{\downarrow_{\scriptsize 2.65}} \\

\hline
\end{tabular}%
}
\end{table*}
To evaluate the robustness of the STAIL framework against the long-tail distribution of the MS-CXR dataset, we conducted experiments across three category ordering scenarios: tail classes appearing at the beginning (Task 0), middle (Task 1), or end (Task 2) of the learning sequence (Table~\ref{table: different category sequential}). The most challenging configuration is the scenario in which tail classes are present in Task 0, where fragile feature spaces constructed from scarce data are highly susceptible to being overwritten by subsequent head-class updates. In this setting, the WA method’s catastrophic forgetting (BWT-AUC -10.30\%) was mitigated by STAIL, which recovered BWT-AUC to -1.12\%, demonstrating that SCB provided textual semantics act as vital anchors for fragile representations.

Similar performance gains were observed when tail classes arrived in later stages. In the situation where tail classes are present in Task 1, TagFex+STAIL achieved a 6.66\% increase in Avg-AUC, facilitating smoother adaptation between tasks of varying data densities. In the scenario where tail classes are present in Task 2, STAIL boosted TagFex performance by nearly 12\% by enhancing the retention of historical knowledge. Ultimately, regardless of the curriculum difficulty or data stream order, these results confirm that STAIL consistently enhances model adaptability and resistance to forgetting through its asymmetric semantic buffering mechanism.
\begin{figure}[width=\linewidth,pos=t]
    \centering
    
    \includegraphics[width=\linewidth]{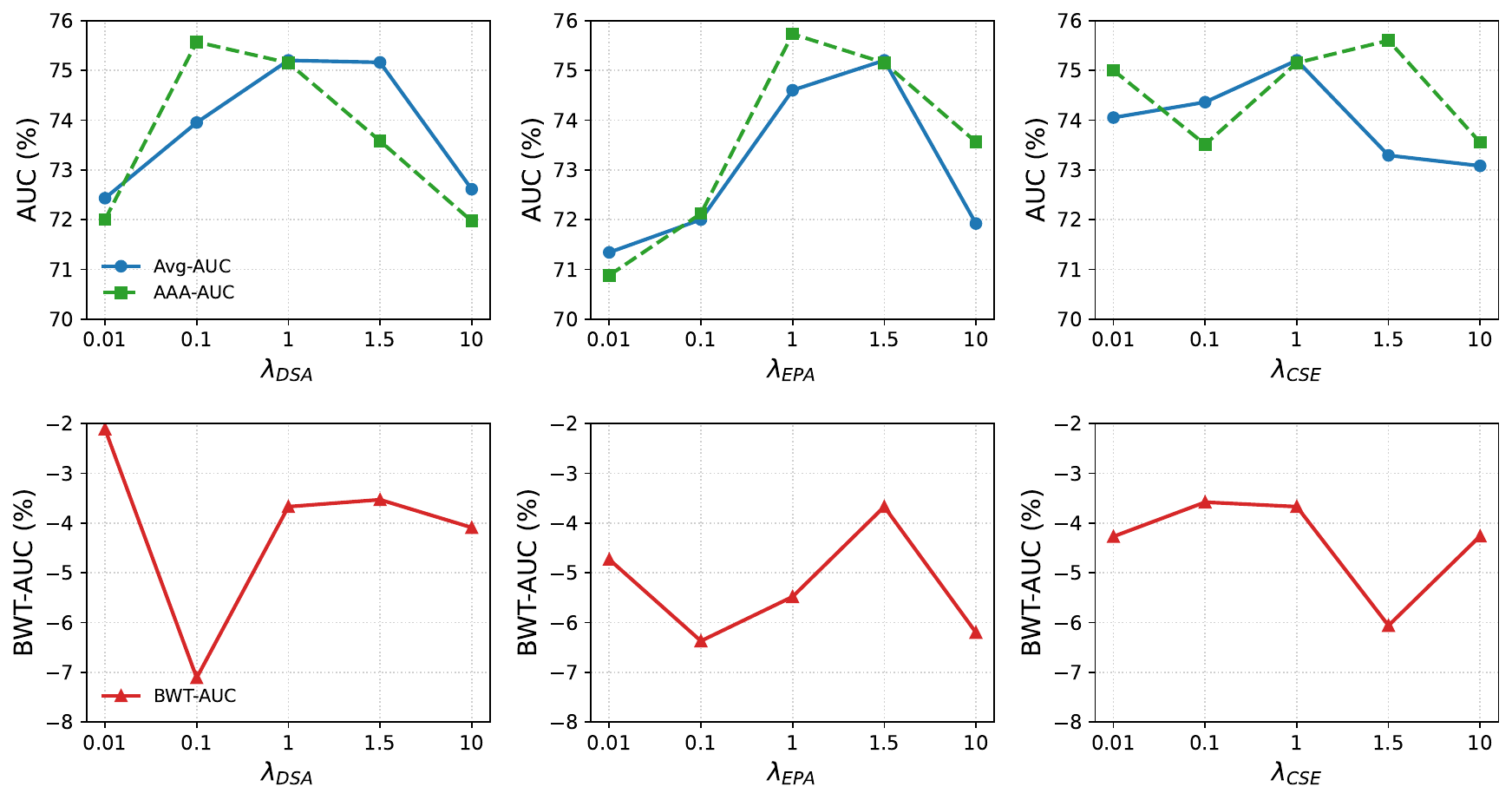}
    
    \caption{Sensitivity analysis on hyperparameters: $\lambda_{DSA}$, $\lambda_{EPA}$, and $\lambda_{CSE}$.}
    \label{fig:hyperparameter}
\end{figure}

\textbf{Hyperparameter Sensitivity Analysis}.
To investigate the sensitivity of the loss function weights in the STAIL framework, we conducted a hyperparameter analysis on the ODIR-5K dataset based on the best-performing TagFex+STAIL baseline. The framework involves three core hyperparameters: $\lambda_{DSA}$, $\lambda_{EPA}$, and $\lambda_{CSE}$. To efficiently navigate the extensive search space, we adopted a step-by-step fixing strategy. Specifically, we sequentially varied one parameter within the search space $\{0.01, 0.1, 1, 1.5, 10\}$ while keeping the other two fixed at their optimal values (i.e., $\lambda_{DSA}=1$, $\lambda_{EPA}=1.5$, and $\lambda_{CSE}=1$).

As illustrated in Fig. \ref{fig:hyperparameter}, the comprehensive performance of the STAIL framework exhibits a classic inverted-U shape with respect to the variations in each hyperparameter, profoundly reflecting the trade-off between stability and plasticity in incremental learning.

It is worth noting that on the ODIR-5K dataset, the aforementioned optimal weight combination ($\lambda_{DSA}=1$, $\lambda_{EPA}=1.5$, $\lambda_{CSE}=1$) is identically applicable to all other baseline methods (Replay, iCaRL, WA) when integrated with STAIL. However, this consistency in weight combination is not observed in other datasets. Given that the US-DATA dataset features significant cross-organ domain shifts and numerous fine-grained categories, while the MS-CXR dataset exhibits an extreme long-tail class distribution, the optimal hyperparameter combinations required by different baseline methods naturally vary when addressing these complex data characteristics. Therefore, in practical clinical deployments, we recommend treating this set of weights from ODIR-5K as a strong initialization for the STAIL framework, which can be fine-tuned according to the specific characteristics of the target medical imaging modality.

\section{CONCLUSION}
\label{cap:conclusion}
We propose STAIL, a biologically inspired framework designed to address catastrophic forgetting and storage inefficiencies in medical class-incremental learning. By shifting the paradigm from pixel-heavy rehearsal to an asymmetric SCB, STAIL achieves a dense reconstruction of historical data manifolds at a fraction of the traditional storage cost. The integration of the LSAM further stabilizes the continuous learning process. By leveraging frozen textual priors as semantic anchors, the LSAM guides visual feature evolution through three complementary loss components. Extensive experiments across heterogeneous medical modalities, including fundus, ultrasound, and X-ray imaging, demonstrate that STAIL consistently outperforms mainstream baselines, even under strict memory constraints. Ultimately, STAIL provides a memory-efficient, cognitively flexible solution, paving the way for the sustainable deployment of robust AI systems in dynamic clinical environments.

\appendix

\section{Supplementary Implementation Details}
\label{Supplementary Implementation Details}

\textbf{Hardware and LLM Quantization}. All experiments were conducted on a server equipped with two NVIDIA A100 (80GB) GPUs. To balance computational performance and memory efficiency, BioMistral-7B was loaded using 4-bit NF4 quantization with double quantization enabled, employing bfloat16 precision for computation.

\textbf{Hyperparameters}. The model's trainable parameters were optimized using the Adam optimizer with $\beta_1=0.9$, $\beta_2=0.999$, and $\epsilon=1\text{e-}8$. The training process consists of an initial learning stage and subsequent incremental learning stages, utilizing distinct hyperparameter schedules. For the initial stage, the model is trained for $160$ epochs. The initial learning rate is set to $0.0001$ and decays by a factor of $0.1$ at the 100th and 140th epochs using a multi-step scheduler; the weight decay is set to $0.0005$. For the incremental stages, each task is trained for $120$ epochs. The learning rate remains $0.0001$ but decays by a factor of $0.1$ at the 70th and 100th epochs, with the weight decay adjusted to $0.0002$. Additionally, for the baseline methods that employ knowledge distillation (iCaRL, WA, and TagFex), the distillation temperature scalar $T$ is set to $2$, and the temperature parameter $\tau$ in the InfoNCE loss is set to $0.07$.

\textbf{Memory Buffer and Exemplar Allocation}. To ensure fair comparison, all competing methods are evaluated under identical memory budgets. Following our proposed SCB strategy, we employ the Herding selection strategy to select representative image exemplars. Specifically, for the ODIR-5K and MS-CXR datasets, we store a fixed number of $20$ image exemplars and $100$ text exemplars per class in the memory buffer $\mathcal{M}$. For the US-DATA dataset, we store $5$ image exemplars and $25$ text exemplars per class.

\section{Theoretical Insights of STAIL}
\label{Theoretical Insights}

This section provides theoretical insights into why STAIL can alleviate catastrophic forgetting in continual medical learning. We analyze the underlying mechanisms of two key components: LLM-derived semantic anchoring and asymmetric semantic memory. The analysis aims to explain how these designs contribute to representation stability and historical knowledge preservation.

\textbf{Semantic Alignment Provides a Stable Representation Reference}. During continual learning, updating the visual encoder with new tasks may cause the learned representation space to deviate from previously acquired semantic structures. STAIL introduces semantic feature anchoring by aligning visual representations with frozen semantic embeddings extracted from a pretrained language model.

Let $z_I$ and $z_S$ denote the visual representation and corresponding semantic representation, respectively. After optimizing the semantic alignment objective, we assume the representation discrepancy is bounded:
$||z_I-z_S||_2\leq \epsilon ,$ where $\epsilon$ represents the residual alignment error.

\textit{Proposition 1 (Semantic Alignment Stabilizes Visual Representations).}
If the semantic alignment error is bounded by $\epsilon$, the distance between two visual representations from different classes is constrained by their corresponding semantic representations:
\begin{equation*}
||z_I^a-z_I^b||
\geq
||z_S^a-z_S^b||-2\epsilon .
\end{equation*}

\textit{Intuition.}
According to the triangle inequality, the deviation between visual and semantic representations limits the distortion of inter-class semantic structures. Therefore, aligning visual features with a fixed semantic space provides a stable reference during incremental updates and reduces arbitrary representation drift.

This observation explains why semantic anchoring can improve the stability of visual representations when learning new clinical tasks.

\textbf{Asymmetric Semantic Memory Enhances Historical Knowledge Coverage}. Conventional rehearsal methods mainly rely on storing a limited number of image exemplars. However, medical images usually exhibit substantial intra-class variation, and a small visual buffer may not fully capture the semantic diversity of previous tasks.

STAIL introduces an asymmetric Semantic Consolidation Buffer (SCB), which maintains a small set of representative image anchors $\mathcal{M}_{img}$ together with a larger collection of semantic descriptions $\mathcal{M}_{text}$. The text memory contains both paired descriptions associated with retained images and supplementary descriptions that provide additional semantic information for historical categories.

\textit{Proposition 2 (Semantic Memory Improves Historical Representation Coverage).}
Under a fixed memory budget, combining visual exemplars with additional semantic descriptions provides a richer representation of historical categories compared with storing images alone.

\textit{Intuition.}
Image exemplars preserve concrete visual characteristics, while textual descriptions provide complementary semantic information about historical classes. Therefore, the asymmetric memory design enlarges the semantic coverage of previous tasks and reduces the dependence on a small number of visual samples. This additional semantic coverage helps the model maintain more complete historical knowledge during incremental updates.

The effectiveness of this mechanism depends on the quality and diversity of available semantic descriptions, which explains why the benefit of semantic consolidation may vary across datasets with different numbers of classes and memory allocation per category.

\section{Pseudo-code}
\label{Pseudo-code}
\begin{algorithm}[H]
\caption{Training Algorithm for STAIL Framework}
\label{alg:stade_refined}
\begin{algorithmic}[1]
\REQUIRE Sequential task datasets $\mathcal{T} = \{\mathcal{D}_1, \mathcal{D}_2, \dots, \mathcal{D}_K\}$, Asymmetric memory budgets $N_{img}, N_{text}$
\ENSURE Trained model parameters $\Theta = \{\theta_{g_I}, \theta_{W_{proj}}, \theta_{C_{Base}}, \{\theta_{g_{proj}^t}\}_{t=1}^K\}$

\STATE \textbf{Initialize Buffer:} Image buffer $\mathcal{M}_{img} \leftarrow \emptyset$, Text buffer $\mathcal{M}_{text} \leftarrow \emptyset$
\STATE \textbf{Initialize Network State:} \textbf{Freeze} LLM text encoder $\theta_{g_{llm}}$

\FOR{each incremental task $t = 1, 2, \dots, K$}
    \STATE \textbf{// 1. Network Status Configuration}
    \IF{$t > 1$}
        \STATE Freeze historical projection heads $\{\theta_{g_{proj}^k}\}_{k=1}^{t-1}$
    \ENDIF
    \STATE \textbf{Initialize} current task-specific projection head $\theta_{g_{proj}^t}$ and set it to be trainable
    \STATE Set visual encoder $\theta_{g_I}$, cross-modal bridge $\theta_{W_{proj}}$, and classifier $\theta_{C_{Base}}$ trainable
    
    \STATE \textbf{// 2. Data Preparation}
    \STATE Construct combined training batch $(I_i, S_i) \sim \mathcal{D}_t \cup (\mathcal{M}_{img}, \mathcal{M}_{text})$
    
    \STATE \textbf{// 3. Training Loop}
    \FOR{each training step}
        \STATE Extract stable semantic features $f_{S_i} = g_{llm}(S_i)$
        \STATE Extract and project visual features $f_{I_i} = W_{proj}g_I(I_i)$         
        \STATE Compute $\mathcal{L}_{Base}^t$, $\mathcal{L}_{DSA}^t$, $\mathcal{L}_{EPA}^t$
        \IF{$t > 1$} \STATE Compute $\mathcal{L}_{CSE}^t$ using frozen past projectors $\{\theta_{g_{proj}^k}\}_{k=1}^{t-1}$
        \ELSE \STATE $\mathcal{L}_{CSE}^t \leftarrow 0$
        \ENDIF
        
        \STATE Compute total loss $\mathcal{L}^t$ by Eq. (\ref{totalloss})
        \STATE Update trainable parameters by minimizing $\mathcal{L}^t$ 
    \ENDFOR
    
    \STATE \textbf{// 4. SCB Update \& Exemplar Construction}
    \FOR{each new class $c \in \mathcal{C}_t$}
        \STATE Select $\mathcal{E}_{img}^c$ (size $\le N_{img}$) via \textit{Herding} strategy
        \STATE Retain $\mathcal{E}_{pair}^c$ for selected images
        \STATE Sample $\mathcal{E}_{extra}^c$ (size $\le N_{text} - N_{img}$) via \textit{Random Sampling} from the difference set $\mathcal{U}_c^t = \mathcal{D}_c^t \setminus \mathcal{E}_{img}^c$
        
        \STATE Update Buffers:  $\mathcal{M}_{img} \leftarrow \mathcal{M}_{img} \cup \mathcal{E}_{img}^c,$
         $\mathcal{M}_{text} \leftarrow \mathcal{M}_{text} \cup (\mathcal{E}_{pair}^c \cup \mathcal{E}_{extra}^c)$
    \ENDFOR
\ENDFOR

\RETURN Final model parameters $\Theta$
\end{algorithmic}
\end{algorithm}



\printcredits

\bibliographystyle{cas-model2-names}

\bibliography{main}

@ARTICLE{zhang2025anti,
  author={Zhang, Yajie and Hu, Yao and Cai, Chengjun and Huang, Yu-An and Huang, Zhi-An and Chen Tan, Kay},
  journal={IEEE Transactions on Neural Networks and Learning Systems}, 
  title={{Anti-Confounding Hashing: Enhancing Radiological Image Retrieval via Debiased Weighting and Counterfactual Reasoning}}, 
  year={2025},
  volume={36},
  number={8},
  pages={15055-15069},
  doi={10.1109/TNNLS.2025.3526760}}

@article{zhang2025causalmixnet,
  title={{CausalMixNet}: A mixed-attention framework for causal intervention in robust medical image diagnosis},
  author={Zhang, Yajie and Huang, Yu-An and Hu, Yao and Liu, Rui and Wu, Jibin and Huang, Zhi-An and Tan, Kay Chen},
  journal={Medical Image Analysis},
  volume={103},
  pages={103581},
  year={2025},
  publisher={Elsevier},
  doi={https://doi.org/10.1016/j.media.2025.103581}
}

@incollection{mccloskey1989catastrophic,
  title={{Catastrophic interference in connectionist networks: The sequential learning problem}},
  author={McCloskey, Michael and Cohen, Neal J},
  booktitle={Psychology of Learning and Motivation},
  volume={24},
  pages={109--165},
  year={1989},
  publisher={Elsevier}
}

@inproceedings{rebuffi2017icarl,
  title={{iCaRL: Incremental Classifier and Representation Learning}},
  author={Rebuffi, Sylvestre-Alvise and Kolesnikov, Alexander and Sperl, Georg and Lampert, Christoph H},
  booktitle={Proceedings of the IEEE Conference on Computer Vision and Pattern Recognition},
  pages={2001--2010},
  year={2017}
}

@INPROCEEDINGS{10888025,
  author={Rahmani, Sana and Chatterjee, Reetam and Etemad, Ali and Hashemi, Javad},
  booktitle={ICASSP 2025 - 2025 IEEE International Conference on Acoustics, Speech and Signal Processing (ICASSP)}, 
  title={{Dynamic Prototype Rehearsal for Continual ECG Arrhythmia Detection}}, 
  year={2025},
  volume={},
  number={},
  pages={1-5},
  doi={10.1109/ICASSP49660.2025.10888025}}

@article{kirkpatrick2017overcoming,
  title={Overcoming catastrophic forgetting in neural networks},
  author={Kirkpatrick, James and Pascanu, Razvan and Rabinowitz, Neil and Veness, Joel and Desjardins, Guillaume and Rusu, Andrei A and Milan, Kieran and Quan, John and Ramalho, Tiago and Grabska-Barwinska, Agnieszka and others},
  journal={Proceedings of the National Academy of Sciences},
  volume={114},
  number={13},
  pages={3521--3526},
  year={2017},
  publisher={National Academy of Sciences}
}

@inproceedings{zenke2017continual,
  title={Continual learning through synaptic intelligence},
  author={Zenke, Friedemann and Poole, Ben and Ganguli, Surya},
  booktitle={International Conference on Machine Learning},
  pages={3987--3995},
  year={2017},
  organization={PMLR}
}

@article{li2017learning,
  title={Learning without forgetting},
  author={Li, Zhizhong and Hoiem, Derek},
  journal={IEEE transactions on Pattern Analysis and Machine Intelligence},
  volume={40},
  number={12},
  pages={2935--2947},
  year={2017},
  publisher={IEEE}
}

@inproceedings{wang2025enhancing,
  title={{Enhancing Continual Learning for Medical Imaging: Efficient Knowledge Transfer and Multi-Disease Prediction}},
  author={Wang, Enzhi and Li, Qicheng and Liu, Di and Yang, Bo},
  booktitle={ICASSP 2025-2025 IEEE International Conference on Acoustics, Speech and Signal Processing (ICASSP)},
  pages={1--5},
  year={2025},
  organization={IEEE}
}

@article{kumaran2016learning,
  title={{What learning systems do intelligent agents need? Complementary learning systems theory updated}},
  author={Kumaran, Dharshan and Hassabis, Demis and McClelland, James L},
  journal={Trends in Cognitive Sciences},
  volume={20},
  number={7},
  pages={512--534},
  year={2016},
  publisher={Elsevier}
}

@article{mcclelland1995there,
  title={Why there are complementary learning systems in the hippocampus and neocortex: insights from the successes and failures of connectionist models of learning and memory.},
  author={McClelland, James L and McNaughton, Bruce L and O'Reilly, Randall C},
  journal={Psychological Review},
  volume={102},
  number={3},
  pages={419},
  year={1995},
  publisher={American Psychological Association}
}

@article{smith2005development,
  title={{The development of embodied cognition: Six lessons from babies}},
  author={Smith, Linda and Gasser, Michael},
  journal={Artificial life},
  volume={11},
  number={1-2},
  pages={13--29},
  year={2005},
  publisher={MIT Press}
}

@article{lake2017building,
  title={Building machines that learn and think like people},
  author={Lake, Brenden M and Ullman, Tomer D and Tenenbaum, Joshua B and Gershman, Samuel J},
  journal={Behavioral and Brain Sciences},
  volume={40},
  pages={e253},
  year={2017},
  publisher={Cambridge University Press}
}

@article{kumari2025continual,
  title={Continual learning in medical image analysis: {A} comprehensive review of recent advancements and future prospects},
  author={Kumari, Pratibha and Chauhan, Joohi and Bozorgpour, Afshin and Huang, Boqiang and Azad, Reza and Merhof, Dorit},
  journal={Medical Image Analysis},
  pages={103730},
  year={2025},
  publisher={Elsevier}
}

@article{wang2024rehearsal,
  title={Rehearsal-free modular and compositional continual learning for language models},
  author={Wang, Mingyang and Adel, Heike and Lange, Lukas and Str{\"o}tgen, Jannik and Sch{\"u}tze, Hinrich},
  journal={arXiv preprint arXiv:2404.00790},
  year={2024}
}

@article{gilboa2017neurobiology,
  title={Neurobiology of schemas and schema-mediated memory},
  author={Gilboa, Asaf and Marlatte, Hannah},
  journal={Trends in Cognitive Sciences},
  volume={21},
  number={8},
  pages={618--631},
  year={2017},
  publisher={Elsevier}
}

@article{mirolli2009language,
  title={Language as a cognitive tool},
  author={Mirolli, Marco and Parisi, Domenico},
  journal={Minds and Machines},
  volume={19},
  number={4},
  pages={517--528},
  year={2009},
  publisher={Springer}
}

@inproceedings{yan2021dynamically,
  title={{DER: Dynamically expandable representation for class incremental learning}},
  author={Yan, Shipeng and Xie, Jiangwei and He, Xuming},
  booktitle={Proceedings of the IEEE/CVF Conference on Computer Vision and Pattern Recognition},
  pages={3014--3023},
  year={2021}
}

@inproceedings{wang2022foster,
  title={{Foster: Feature boosting and compression for class-incremental learning}},
  author={Wang, Fu-Yun and Zhou, Da-Wei and Ye, Han-Jia and Zhan, De-Chuan},
  booktitle={European Conference on Computer Vision},
  pages={398--414},
  year={2022},
  organization={Springer}
}

@inproceedings{zhoumodel,
  title={{A Model or 603 Exemplars: Towards Memory-Efficient Class-Incremental Learning}},
  author={Zhou, Da-Wei and Wang, Qi-Wei and Ye, Han-Jia and Zhan, De-Chuan},
  booktitle={The Eleventh International Conference on Learning Representations},
  year = {2022}
}

@inproceedings{wang2022beef,
  title={{Beef: Bi-compatible class-incremental learning via energy-based expansion and fusion}},
  author={Wang, Fu-Yun and Zhou, Da-Wei and Liu, Liu and Ye, Han-Jia and Bian, Yatao and Zhan, De-Chuan and Zhao, Peilin},
  booktitle={The eleventh International Conference on Learning Representations},
  year={2022}
}

@inproceedings{zheng2025task,
  title={{Task-Agnostic Guided Feature Expansion for Class-Incremental Learning}},
  author={Zheng, Bowen and Zhou, Da-Wei and Ye, Han-Jia and Zhan, De-Chuan},
  booktitle={Proceedings of the Computer Vision and Pattern Recognition Conference},
  pages={10099--10109},
  year={2025}
}

@inproceedings{radford2021CLIP,
  title={{Learning Transferable Visual Models From Natural Language Supervision}},
  author={Radford, Alec and Kim, Jong Wook and Hallacy, Chris and Ramesh, Aditya and Goh, Gabriel and Agarwal, Sandhini and Sastry, Girish and Askell, Amanda and Mishkin, Pamela and Clark, Jack and others},
  booktitle={International Conference on Machine Learning},
  pages={8748--8763},
  year={2021},
  organization={PMLR}
}

@inproceedings{wang2022L2P,
  title={{Learning to Prompt for Continual Learning}},
  author={Wang, Zifeng and Zhang, Zizhao and Lee, Chen-Yu and Zhang, Han and Sun, Ruoxi and Ren, Xiaoqi and Su, Guolong and Perot, Vincent and Dy, Jennifer and Pfister, Tomas},
  booktitle={Proceedings of the IEEE/CVF Conference on Computer Vision and Pattern Recognition},
  pages={139--149},
  year={2022}
}

@inproceedings{wu2024meta,
  title={{Meta continual learning revisited: Implicitly enhancing online hessian approximation via variance reduction}},
  author={Wu, Yichen and Huang, Long-Kai and Wang, Renzhen and Meng, Deyu and Wei, Ying},
  booktitle={The Twelfth International Conference on Learning Representations},
  volume={2},
  year={2024}
}

@article{caccia2021new,
  title={New insights on reducing abrupt representation change in online continual learning},
  author={Caccia, Lucas and Aljundi, Rahaf and Asadi, Nader and Tuytelaars, Tinne and Pineau, Joelle and Belilovsky, Eugene},
  journal={arXiv preprint arXiv:2104.05025},
  year={2021}
}

@inproceedings{zhao2020maintaining,
  title={Maintaining discrimination and fairness in class incremental learning},
  author={Zhao, Bowen and Xiao, Xi and Gan, Guojun and Zhang, Bin and Xia, Shu-Tao},
  booktitle={Proceedings of the IEEE/CVF Conference on Computer Vision and Pattern Recognition},
  pages={13208--13217},
  year={2020}
}

@article{zhou2023pycil,
    author = {Da-Wei Zhou and Fu-Yun Wang and Han-Jia Ye and De-Chuan Zhan},
    title = {{PyCIL}: a Python toolbox for class-incremental learning},
    journal = {SCIENCE CHINA Information Sciences},
    year = {2023},
    volume = {66},
    number = {9},
    pages = {197101},
    doi = {https://doi.org/10.1007/s11432-022-3600-y}
  }

@inproceedings{he2016deep,
  title={Deep residual learning for image recognition},
  author={He, Kaiming and Zhang, Xiangyu and Ren, Shaoqing and Sun, Jian},
  booktitle={Proceedings of the IEEE Conference on Computer Vision and Pattern Recognition},
  pages={770--778},
  year={2016}
}

@inproceedings{welling2009herding,
  title={Herding dynamical weights to learn},
  author={Welling, Max},
  booktitle={Proceedings of the 26th Annual International Conference on Machine Learning},
  pages={1121--1128},
  year={2009}
}

@article{labrak2024biomistral,
  title={{Biomistral: A collection of open-source pretrained large language models for medical domains}},
  author={Labrak, Yanis and Bazoge, Adrien and Morin, Emmanuel and Gourraud, Pierre-Antoine and Rouvier, Mickael and Dufour, Richard},
  journal={arXiv preprint arXiv:2402.10373},
  year={2024}
}

@article{lin2022beyond,
  title={{Beyond not-forgetting: Continual learning with backward knowledge transfer}},
  author={Lin, Sen and Yang, Li and Fan, Deliang and Zhang, Junshan},
  journal={Advances in Neural Information Processing Systems},
  volume={35},
  pages={16165--16177},
  year={2022}
}

@inproceedings{liturning,
  title={{Turning the Tables: Enabling Backward Transfer via Causal-Aware LoRA in Continual Learning}},
  author={Li, Chaoyang and Ye, Runze and Qin, Jianyang and Cui, Jinhao and Wang, Lingzhi and Hu, Ning and Liao, Qing},
  booktitle={The Thirty-ninth Annual Conference on Neural Information Processing Systems},
  year={2025}
}

@article{chen2023meditron,
  title={{Meditron-70b: Scaling medical pretraining for large language models}},
  author={Chen, Zeming and Cano, Alejandro Hern{\'a}ndez and Romanou, Angelika and Bonnet, Antoine and Matoba, Kyle and Salvi, Francesco and Pagliardini, Matteo and Fan, Simin and K{\"o}pf, Andreas and Mohtashami, Amirkeivan and others},
  journal={arXiv preprint arXiv:2311.16079},
  year={2023}
}

@misc{qwen2.5,
    title = {{Qwen2.5: A Party of Foundation Models}},
    url = {https://qwenlm.github.io/blog/qwen2.5/},
    author = {Qwen Team},
    month = {September},
    year = {2024}
}

@article{PhysioNet-ms-cxr-1.1.0,
  author = {Boecking, Benedikt and Usuyama, Naoto and Bannur, Shruthi and {Coelho de Castro}, Daniel and Schwaighofer, Anton and Hyland, Stephanie and Sharma, Harshita and Wetscherek, Maria Teodora and Naumann, Tristan and Nori, Aditya and {Alvarez Valle}, Javier and Poon, Hoifung and Oktay, Ozan},
  title = {{MS-CXR: Making the Most of Text Semantics to Improve Biomedical Vision-Language Processing}},
  journal = {{PhysioNet}},
  year = {2024},
  month = nov,
  note = {Version 1.1.0},
  doi = {10.13026/9g2z-jg61},
  url = {https://doi.org/10.13026/9g2z-jg61}
}

@misc{odir2019,
  title = {{International Competition on Ocular Disease Intelligent Recognition}},
  author = {Peking University and Shanggong Medical Technology Co., Ltd.},
  year = {2019},
  howpublished = {\url{https://odir2019.grand-challenge.org}}
}

@article{li2024ultrasound,
  title={Ultrasound report generation with cross-modality feature alignment via unsupervised guidance},
  author={Li, Jun and Su, Tongkun and Zhao, Baoliang and Lv, Faqin and Wang, Qiong and Navab, Nassir and Hu, Ying and Jiang, Zhongliang},
  journal={IEEE Transactions on Medical Imaging},
  volume={44},
  number={1},
  pages={19--30},
  year={2024},
  publisher={IEEE}
}

@article{wu2024pmc,
  title={{PMC-LLaMA}: toward building open-source language models for medicine},
  author={Wu, Chaoyi and Lin, Weixiong and Zhang, Xiaoman and Zhang, Ya and Xie, Weidi and Wang, Yanfeng},
  journal={Journal of the American Medical Informatics Association},
  volume={31},
  number={9},
  pages={1833--1843},
  year={2024},
  publisher={Oxford University Press}
}

@article{grattafiori2024llama,
  title={The llama 3 herd of models},
  author={Grattafiori, Aaron and Dubey, Abhimanyu and Jauhri, Abhinav and Pandey, Abhinav and Kadian, Abhishek and Al-Dahle, Ahmad and Letman, Aiesha and Mathur, Akhil and Schelten, Alan and Vaughan, Alex and others},
  journal={arXiv preprint arXiv:2407.21783},
  year={2024}
}

@article{qian2025cpsr,
  title={{CPSR-CLIP: Conditional Prompt-Induced Style Reconstruction for Zero-Shot Domain Adaptation}},
  author={Qian, Jiayu and Lu, Yuwu and Xie, Wuyuan and Lai, Zhihui and Wang, Miaohui and Li, Xuelong},
  journal={IEEE Transactions on Multimedia},
  year={2025},
  publisher={IEEE}
}

@inproceedings{selvaraju2017grad,
  title={{Grad-CAM: Visual Explanations from Deep Networks via Gradient-based Localization}},
  author={Selvaraju, Ramprasaath R and Cogswell, Michael and Das, Abhishek and Vedantam, Ramakrishna and Parikh, Devi and Batra, Dhruv},
  booktitle={Proceedings of the IEEE International Conference on Computer Vision},
  pages={618--626},
  year={2017}
}

@article{li2025progressive,
  title={{Progressive Distillation for Incremental Learning in Corneal Confocal Microscopy Segmentation}},
  author={Li, Hongshuo and Ma, Baikai and Mou, Lei and Liu, Yonghuai and Zheng, Qinxiang and Qi, Hong and Zhao, Yitian},
  journal={IEEE Transactions on Medical Imaging},
  year={2025},
  publisher={IEEE}
}

@article{wang2025cross,
  title={{Cross-Domain Invariant Feature Absorption and Domain-Specific Feature Retention for Domain Incremental Chest X-Ray Classification}},
  author={Wang, Mengchu and He, Yuhang and Peng, Lin and Song, Xiang and Dong, Songlin and Gong, Yihong},
  journal={IEEE Transactions on Medical Imaging},
  volume={44},
  number={5},
  pages={2041--2055},
  year={2025},
  publisher={IEEE}
}

@article{bayasi2024gc,
  title={{${GC}^2$: Generalizable Continual Classification of Medical Images}},
  author={Bayasi, Nourhan and Hamarneh, Ghassan and Garbi, Rafeef},
  journal={IEEE Transactions on Medical Imaging},
  volume={43},
  number={11},
  pages={3767--3779},
  year={2024},
  publisher={IEEE}
}

@article{thandiackal2024multi,
  title={{Multi-Scale Feature Alignment for Continual Learning of Unlabeled Domains}},
  author={Thandiackal, Kevin and Piccinelli, Luigi and Gupta, Rajarsi and Pati, Pushpak and Goksel, Orcun},
  journal={IEEE Transactions on Medical Imaging},
  volume={43},
  number={7},
  pages={2599--2609},
  year={2024},
  publisher={IEEE}
}

@article{huang2026scbit,
  title={{scBIT: Integrating Single-cell Transcriptomic Data into fMRI-based Prediction for Alzheimer's Disease Diagnosis}},
  author={Huang, Yu-An and Hu, Yao and Li, Yue-Chao and Cao, Xiyue and Li, Xinyuan and Tan, Kay Chen and You, Zhu-Hong and Huang, Zhi-An},
  journal={IEEE Transactions on Medical Imaging},
  year={2026},
  publisher={IEEE}
}

@article{wang2025rethinking,
  title={{Rethinking Class-Incremental Learning From a Dynamic Imbalanced Learning Perspective}},
  author={Wang, Leyuan and Xiang, Liuyu and Wang, Yunlong and Wu, Huijia and Yang, Huafeng and Liu, Jingqian and He, Zhaofeng},
  journal={IEEE Transactions on Multimedia},
  volume={28},
  pages={825--836},
  year={2025},
  publisher={IEEE}
}

@ARTICLE{11329432,
  author={Chen, Bingzhi and Chen, Zhiming and Cai, Sudong and Fang, Xiaozhao and Bennamoun, Mohammed and Xie, Shengli},
  journal={IEEE Transactions on Multimedia}, 
  title={{Toward Bidirectional Adaptability for Few-Shot Class-Incremental Learning With Forward-Backward Knowledge Transfer}}, 
  year={2026},
  volume={28},
  number={},
  pages={2337-2351},
  doi={10.1109/TMM.2026.3651015}}

@article{wang2025dual,
  title={{Dual-Attention Transformers for Class-Incremental Learning: A Tale of Two Memories}},
  author={Wang, Shaofan and Wang, Weixing and Sun, Yanfeng and Wang, Zhiyong and Wang, Boyue and Yin, Baocai},
  journal={IEEE Transactions on Multimedia},
  year={2025},
  volume={27},
  pages={8763-8775},
  publisher={IEEE},
  doi={10.1109/TMM.2025.3607800}
}

@article{cheng2025distribution,
  title={{Distribution-Level Memory Recall for Continual Learning: Preserving Knowledge and Avoiding Confusion}},
  author={Cheng, Shaoxu and Geng, Kanglei and He, Chiyuan and Qiu, Zihuan and Xu, Linfeng and Qiu, Heqian and Wang, Lanxiao and Wu, Qingbo and Meng, Fanman and Li, Hongliang},
  journal={IEEE Transactions on Multimedia},
  volume={27},
  pages={4151--4166},
  year={2025},
  publisher={IEEE}
}

@inproceedings{wang2022dualprompt,
  title={{DualPrompt: Complementary Prompting for Rehearsal-free Continual Learning}},
  author={Wang, Zifeng and Zhang, Zizhao and Ebrahimi, Sayna and Sun, Ruoxi and Zhang, Han and Lee, Chen-Yu and Ren, Xiaoqi and Su, Guolong and Perot, Vincent and Dy, Jennifer and others},
  booktitle={European Conference on Computer Vision},
  pages={631--648},
  year={2022},
  organization={Springer}
}

@inproceedings{boecking2022making,
  title={{Making the Most of Text Semantics to Improve Biomedical Vision--Language Processing}},
  author={Boecking, Benedikt and Usuyama, Naoto and Bannur, Shruthi and Castro, Daniel C and Schwaighofer, Anton and Hyland, Stephanie and Wetscherek, Maria and Naumann, Tristan and Nori, Aditya and Alvarez-Valle, Javier and others},
  booktitle={European Conference on Computer Vision},
  pages={1--21},
  year={2022},
  organization={Springer}
}

@article{ayromlou2024ccsi,
  title={{CCSI: Continual Class-Specific Impression for data-free class incremental learning}},
  author={Ayromlou, Sana and Tsang, Teresa and Abolmaesumi, Purang and Li, Xiaoxiao},
  journal={Medical Image Analysis},
  volume={97},
  pages={103239},
  year={2024},
  publisher={Elsevier}
}

@article{BUI2026104235,
title = {Welcome new doctor: Continual learning with expert consultation and autoregressive inference for whole slide image analysis},
journal = {Medical Image Analysis},
volume = {114},
pages = {104235},
year = {2026},
issn = {1361-8415},
doi = {https://doi.org/10.1016/j.media.2026.104235},
url = {https://www.sciencedirect.com/science/article/pii/S136184152600304X},
author = {Doanh C. Bui and Jin Tae Kwak}
}

@article{bayasi2025biaspruner,
  title={BiasPruner: Mitigating bias transfer in continual learning for fair medical image analysis},
  author={Bayasi, Nourhan and Fayyad, Jamil and Bissoto, Alceu and Hamarneh, Ghassan and Garbi, Rafeef},
  journal={Medical image analysis},
  pages={103764},
  year={2025},
  publisher={Elsevier}
}

@inproceedings{shui2025large,
  title={{Large-scale and Fine-grained Vision-Language Pre-training for Enhanced CT Image Understanding}},
  author={Shui, Zhongyi and Zhang, Jianpeng and Cao, Weiwei and Wang, Sinuo and Guo, Ruizhe and Lu, Le and Yang, Lin and Ye, Xianghua and Liang, Tingbo and Zhang, Qi and others},
  booktitle={International Conference on Learning Representations},
  volume={2025},
  pages={24094--24107},
  year={2025}
}





\end{document}